\documentclass[USenglish,oneside,twocolumn]{article}

\usepackage[big]{dgruyter_NEW}

\usepackage{microtype}
\usepackage{amsfonts}
\usepackage[table,xcdraw]{xcolor}
\usepackage{booktabs}
\usepackage{graphicx}
\usepackage{subfig}
\usepackage{multirow}
\usepackage{csquotes}
\usepackage{algorithmic}
\usepackage[linesnumbered, ruled, vlined]{algorithm2e}

\usepackage{amsmath}
\usepackage{tikz}
\tikzstyle{block} = [draw, fill=none, rectangle, minimum height=3em, minimum width=6em]
\newcommand{\vapprox}{\rotatebox{90}{$\approx$}}
\usetikzlibrary{patterns,decorations.pathmorphing,decorations.pathreplacing}


\newcommand{\zhigang}[1]{{\color{black}#1}}
\newcommand{\gio}[1]{{\color{black}#1}}
\newcommand{\shakila}[1]{{\color{black}#1}}

\newif\ifcomment
\commenttrue
\ifcomment

    \newcounter{GTNumberOfComments}
    \stepcounter{GTNumberOfComments}
    \newcommand{\gtnote}[1]{\textcolor{violet}{\small \bf [GT: \#\arabic{GTNumberOfComments}\stepcounter{GTNumberOfComments}: #1]}}

    \newcommand{\NOTE}[1]
    {
      {\footnotesize\it
        \begin{center}
          \begin{tabular}{|c|}
           \hline
            \parbox{0.85\columnwidth}{
              \medskip
              #1
              \medskip} \\
            \hline
          \end{tabular}
        \end{center}
        }
    }
\else
      \newcommand\gtnote[1]{}
    \newcommand\NOTE[1]{}
\fi

\DOI{foobar}

\cclogo{\includegraphics{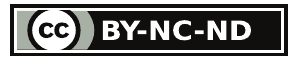}}
  
\begin{document}

   \author*[1]{Shakila Mahjabin Tonni}

   \author[2]{Dinusha Vatsalan}
  
  \author[3]{Farhad Farokhi}

   \author[4]{Dali Kaafar}
  
   \author[5]{Zhigang Lu}
  
   \author[6]{Gioacchino Tangari}

  \affil[1]{Macquarie University, E-mail: shakila.tonni@mq.edu.au}

  \affil[3]{University of Melbourne, E-mail: farhad.farokhi@unimelb.edu.au}

  \affil[2]{Data61, CSIRO, E-mail: dinusha.vatsalan@data61.csiro.au}

  \affil[4]{Macquarie University, E-mail: dali.kaafar@mq.edu.au}
  
  \affil[5]{Macquarie University, E-mail: zhigang.lu@mq.edu.au}
  
  \affil[6]{Macquarie University, E-mail: gioacchino.tangari@mq.edu.au}

\title{\huge Data and Model Dependencies of Membership Inference Attacks}

  \runningtitle{Data and Model Dependencies of Membership Inference Attacks}


\begin{abstract}
{Machine learning (ML) models have been shown to be vulnerable to Membership Inference Attacks (MIA), which infer the membership of a given data point in the target dataset by observing the prediction output of the ML model. 
While the key factors for the success of MIA have not yet been fully understood, existing defense mechanisms such as using L2 regularization~\cite{10shokri2017membership} and dropout layers~\cite{salem2018ml} take only the model's overfitting property into consideration. In this paper, we provide an empirical analysis of the impact of both the data and ML model properties on the vulnerability of ML techniques to MIA. 
Our results reveal the relationship between MIA accuracy and properties of the dataset and training model in use. In particular, we show that the size of shadow dataset, the class and feature balance and the entropy of the target dataset, the configurations and fairness of the training model are the most influential factors. Based on those experimental findings, we conclude that along with model overfitting, multiple properties jointly contribute to MIA success instead of any single 
property. Building on our experimental findings, we propose using those data and model properties as regularizers to protect ML models against MIA. Our results show that the proposed defense mechanisms can reduce the MIA accuracy by up to 25\% without sacrificing the ML model prediction utility.}
\end{abstract}


 \journalname{}
\DOI{Editor to enter DOI}
  \startpage{1}
  \received{..}
  \revised{..}
  \accepted{..}

  \journalyear{..}
  \journalvolume{..}
  \journalissue{..}
 
\maketitle

\section{Introduction}
\label{sec:intro}

Machine Learning as a Service (MLaaS) are fully managed services providing data scientists and developers, and more generally data-driven organisations, with the ability to build, train and deploy machine learning models. Third-party platforms are increasingly deploying publicly accessible query interfaces (Amazon, Google, and Microsoft) allowing users to train models on (potentially sensitive) data while charging access to the models on a pay-per-query basis.
Although promising, these "open and publicly available query-based" platforms face several security and privacy challenges, in particular when considering adversarial attacks probing the outcome of the model to gain knowledge of the model structure or to infer additional information about some users' records.



 
Amongst those threats, Membership Inference Attacks (MIA) as presented in a seminal work by Shokri et al.~\cite{10shokri2017membership}, where an adversary seeks to infer whether a given data record belongs to a model's training dataset given a (black-box) access to the ML model, is one of the most prominent privacy challenges.

While some prior researches indicate the data-dependency of the MIA~\cite{truex2019effects, 18truex2018towards} and have attributed the success of membership inference attacks to some ML model properties~\cite{10shokri2017membership, nasr2018machine, yeom2018privacy} including the observation that machine learning algorithms tend to return higher confidence scores for examples that they have seen (i.e., records in the training dataset) versus those that they encounter for the first time (e.g. overfitting, choice of hyperparameters), understanding the main reasons behind successful MIAs and why some ML models might be more immune than others is key to designing suitable defense mechanisms.


In this paper, we provide an empirical analysis to capture the relation between MIA success, as studied by ~\cite{10shokri2017membership} and a wide range of data and ML model properties. We leverage our analysis to investigate how the relevant properties can be used to improve the ML model's resistance to MIA without sacrificing its prediction accuracy.   
The paper makes the following contributions: 

\begin{itemize}
    \item We experimentally verify the impacts of different data properties with MIA success on three well-known real-world datasets. Specifically, we observe that the size of the shadow dataset used by the attacker and the feature balance of the target training dataset have positive impact on MIA accuracy, i.e. MIA accuracy increases with an increasing value of a property, while the class balance and the entropy of the target training dataset entropy have negative impacts. 
    Additionally, the number of features used for training the target model does not show explicit impact on MIA accuracy. 
    
    \item {We further explore the relation between ML model properties and MIA accuracy. Our analysis shows that increasing the depth of ML models and size (number of nodes of each layer) achieves more accurate membership inference attacks. It also reveals that trained models with fairer predictive behaviors across records and subgroups exhibit higher resilience against MIA. Furthermore, our study shows that MIA accuracy is heavily dependent on the combination of target and shadow models, and more effective MIAs can be achieved by combining the use of different types of shadow models.}
    
    \item{Building on our experimental findings, we propose the use of influential model properties on MIA as regularizers for the model training. In particular, we investigate regularizers based on model's fairness and on the mutual information between dataset records and model parameters.
    Our evaluation results show that training ML models with the proposed regularizers can reduce MIA accuracy by up to 25\%. Interestingly, this is achieved while increasing the model performance compared to both models using no regularization and models using L1 or L2 regularizers.}
\end{itemize}

The remainder of this paper is structured as follows. 
Section~\ref{sec:background} introduces related work.
In Section~\ref{sec:outline}, we describe our methodology and the experimental setup. We discuss the results from the experimental study in Section~\ref{sec:result}. We propose a new defense method to improve ML models' resilience against MIA in Section~\ref{sec:defense}. We conclude and provide remarks on future work in Section~\ref{sec:conclusion}.


\section{Overview \& Related Work}
\label{sec:background}
We briefly overview the black-box Membership Inference Attacks (MIA) and discuss the related work in analyzing MIA and proposed defense approaches.

\subsection{Membership Inference Attack (MIA)}
\label{sec:lit}

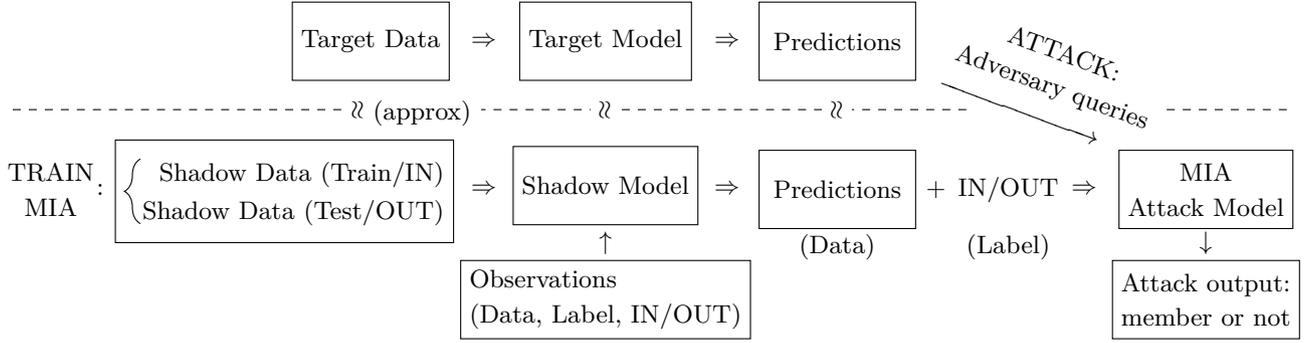
\begin{figure*}[th!]
    \centering
    \captionsetup{justification=centering}
    \begin{tikzpicture}[xscale=0.5, yscale=0.5]
        \draw [dashed] (-11.5,-2) -- (22.2,-2);
        \node [align=center] at (-10.5,-4.1) {TRAIN \\ MIA};
        \node at (-9.2,-4.1) {:};
        \node [block, left] at (0,-0.2) {Target Data};
        \node [below, fill=white] at (-1,-1.5) {$\vapprox$ (approx)};
        \draw [fill=none] (-8.8,-2.8) rectangle (0.1,-5.5);
        \draw [decorate,decoration={brace,amplitude=6pt},xshift=-4pt,yshift=0pt] (-8,-5) -- (-8,-3.3);
        \node [left] at (0.1,-3.7) {Shadow Data (Train/IN)};
        \node [left] at (0.1,-4.7) {Shadow Data (Test/OUT)};
        \node at (0.9,-0.2) {$\Rightarrow$};
        \node at (0.9,-4.1) {$\Rightarrow$};
        \node [block] at (4.1,-0.2) {Target Model};
        \node [below, fill=white] at (4.1,-1.5) {$\vapprox$};
        \node [block] at (4.1,-4) {Shadow Model};
        \node at (4.1,-5.5) {$\uparrow$};
        \node [block, align=left] at (4.1,-7) {Observations \\ (Data, Label, IN/OUT)};
        \node at (7.3,-0.2) {$\Rightarrow$};
        \node at (7.3,-4.1) {$\Rightarrow$};
        \node [block] at (10.2,-0.2) {Predictions};
        \node [below, fill=white] at (10.2,-1.5) {$\vapprox$};
        \node [block] at (10.2,-4.1) {Predictions};
        \node at (10.2,-5.6) {(Data)};
        \node at (12.8,-4.1) {$+$};
        \node at (14.7,-4.1) {IN/OUT};
        \node at (14.7,-5.6) {(Label)};
        \node at (16.7,-4.1) {$\Rightarrow$};
        \node [block, align=center] at (19.9,-4.1) {MIA \\ Attack Model};
        \node [rotate=340, fill=white] at (15,-2) {$\xrightarrow{\hspace*{2cm}}$};
        \node [rotate=340, align=center, fill=white] at (16,-1) {ATTACK: \\ Adversary queries};
        \node at (19.9,-5.5) {$\downarrow$};
        \node [block, align=left] at (19.9,-7) {Attack output: \\ member or not};
    \end{tikzpicture}
    \caption{Training and Attacking of MIA (using the shadow model approach~\cite{10shokri2017membership}) attack model. Dataset IN/OUT is used for training/testing the shadow model, then the IN/OUT are served as labels in attack model training.}
    \label{fig:mia}
\end{figure*}

Typically, a black-box MIA \cite{10shokri2017membership, 18truex2018towards} aims to learn whether a record is part of the private training data which is accessible by permitted parties without the details about the training data and training model. In this paper, we instantiate MIA by the shadow model approach taken by Shokri et al.~\cite{10shokri2017membership}. Their approach relies on three key assumptions about the attacker's capabilities. Firstly, the attacker can acquire the model's prediction vector on any data record. Secondly, the attacker knows the distribution of the target training dataset. Thirdly, the attacker has the information about the target model's structure.

The MIA shadow-model approach, represented in Figure~\ref{fig:mia}, is based on training two types of models: shadow model and attack model. 
The purpose of a shadow model is to imitate the target model's behaviour and produce outputs similar to it. 
The attack model is a binary classifier that categorizes records into member and non-member classes. 
The attack model is trained on the prediction vectors and the class labels obtained from the shadow model and tested against 
the class labels of the target model (truth data/ground truth). 

Further more, Salem et al.~\cite{salem2018ml} proposed three relaxations of the shadow model to reduce the cost of performing MIA. Such relaxations gradually removes the number of shadow models and the dependency on the background knowledge about the model structure and the distribution of training dataset until none of them is using. It finally ends up with a novel MIA inferring the membership by measuring the statistical information of the target model's posterior without the shadow model.
 
MIA has been shown to be successful on various ML models such as Artificial Neural Networks (ANN)~\cite{10shokri2017membership, salem2018ml}, Generative Adversarial Networks (GAN)~\cite{17hayes2019logan, chen2019gan, hilprecht2019monte} and differentially private models~\cite{12rahman2018membership} in a range of domains such as social media~\cite{liu2019socinf}, health~\cite{backes2016membership}, sequence-to-sequence video captioning~\cite{hisamoto2019membership} and user mobility~\cite{pyrgelis2017knock}. 

\subsection{MIA Analysis and Defense Approaches}[t]
\label{sec:rw2}
\zhigang{\textbf{Understanding the reasons behind MIA.}} 
Several studies consider the target model's overfitting to be a primary reason behind MIA's success ~\cite{10shokri2017membership, irolla2019demystifying, yeom2018privacy, 17hayes2019logan}. For better understanding of MIA, Truex et al.~\cite{truex2019effects, 18truex2018towards} empirically studied the increased vulnerability of the target model against MIA introduced by the minority records in a skewed dataset as the performance of the model is not uniform across all classes, hence the attacker. The analysis in ~\cite{truex2019effects} also reveals that attack models can be transferable from one target model type to another, which allows attacker to launch effective MIAs even when the exact target model configuration is not known. 


The study in \cite{18truex2018towards} finds a correlation between the target model's vulnerability to MIA and dataset characteristics such as the feature distribution and the number of classes. An increased risk of membership inference attacks attributed to a higher number of classes in case of multi-class datasets is also reported in~\cite{10shokri2017membership}. 

To further investigate the resilience of ML models to membership inference attacks, the authors of \cite{yaghini2019disparate} introduce the concept of   \textit{MIA-indistinguishability}. Perfect MIA-indistinguishability indicates that, for a given prediction output of the ML model, there are equal probabilities that the input record is and is not part of the training dataset. A model whose prediction behavior is close to perfect MIA-indistinguishability has therefore strong defenses against MIAs.



\textbf{Defense against MIA.} ~A successful MIA exploits a model's tendency to yield higher confidence value when encountering data that they are trained on (members) than the others. 

The property of a model to overfitting towards its training data makes it vulnerable to MIA \cite{yeom2018privacy}. Thus, existing defenses against MIA attempt to reduce overfitting of a model by applying regularizers like L2-regularizer \cite{10shokri2017membership} that generalize a model's prediction or by adding Dropout layers to the model \cite{19srivastava2014dropout} that ignore a few neurons in each iteration of training to avoid high training accuracy. Nasr et al. \cite{ nasr2018machine} proposed a min-max game-theoretic defense method using the highest possible MIA accuracy as adversarial regularization to decrease the target model's prison loss while increasing privacy against MIA.

Recent research~\cite{long2018understanding} revealed that overfitting is a necessary but not a sufficient condition for a successful MIA, by showing that highly successful MIAs can be obtained even for well-generalized models. This poses the need for new defense strategies against MIA, different from the existing overfitting-reduction approaches. One of the alternative defense strategies is to use differential privacy \cite{21mcsherry2009differentially, pyrgelis2017knock, 17hayes2019logan, bassily2014private}. Differential privacy can protect the training dataset privacy using data perturbations at the cost of model accuracy reductions. However, the choice of the level of privacy (privacy budget $\epsilon$) is still an open issue, since the model accuracy may be heavily affected even for modest privacy budgets~\cite{truex2019effects}. Furthermore, authors of~\cite{pyrgelis2017knock} showed that defense mechanisms based on differential privacy are not always effective, especially when an attacker is able to mimic the behavior of the perturbation.


\section{Methodology}\label{sec:outline}

\zhigang{The goal of this study is to analyze the effect of a wide range of properties of dataset and ML model on MIA accuracy. In this section, we introduce the explored properties and describe our experimental setup.}


\subsection{Explored Properties and Their Measures}
\label{sec:data}
As the target model for MIA, we consider a generic ML classifier trained with dataset $D(X,y)$, where each record $x_{i} \in X$ is a vector of features and has a label $y_{i} \in y$. Each record $x_i \in X$ contains a set of feature values $x_i.a_j$ for each feature $a_j\in A$. We denote the set of the distinct feature values for a single feature $a_j$ as $C_j = \{\cup_{x_i\in X}~{x_i.a_j}\}_{\neq}$.

\textbf{Data Properties.} 
Recent research has shown that the performance of MIA is generally data dependent, with significantly different attack success rates measured on different datasets~\cite{truex2019effects, 18truex2018towards}. 
%
In this study, we dissect the impact of the dataset on MIA focusing on key data properties such as the data size, balance in the classes and features, number of selected features, and entropy of the dataset. These are detailed below.

    
\textit{The balance in classes. }
    ~To investigate the impact of the class balance on MIA performance, we assume the target model to be a binary classifier, such that the balance between classes $y\in \{0,1\}$ is given by $P[y=0] / P[y=1]$ and measured as the frequency ratio between the two classes over the dataset. For simplicity, in the next sections we denote the class balance as the percentage of one of the class labels.

    

\textit{The balance in features.}
    ~For a single feature $a_j \in A$, feature balance is the probability ratio of one versus all the other feature values in $C_j$.
    Similarly, for the given dataset $D(X, y)$, the set of the distinct feature vectors for the feature set $A$ is $C = \{\cup_{x_{i} \in X}x_{i}\}_{\neq}$, then the feature balance can be measured as the ratio between one unique value of feature vector against others $ P[x_i = c | c\in C] / P[x_i \neq c | c\in C]$. 
    
  
\textit{The entropy of the training dataset. }
    ~Entropy of the overall dataset $H_D$ can be measured by taking the mean entropy over $n$ number of features:
    \begin{align}\label{eq:entropy}
            H_D = \frac{1}{n} \sum_{j} H[a_j],
    \end{align}
    \shakila{where $a_j \in A$ are the features of the dataset $D$. The entropy of each feature $H[a_j]$ could be computed using Shannon's entropy formula~\cite{entropy_2019} as below:
    \begin{equation}
       H[a_j] = - \sum_{c \in C_j} P_{c} \log P_{c},
    \end{equation}
    where $P_c$ is the probability of each feature value $c \in C_j$. 
    }



\textbf{Model Properties.}
Our study analyzes a variety of properties concerning the ML model and its prediction behavior. 
Recent studies~\cite{nasr2018machine, 10shokri2017membership, truex2019effects} have identified overfitting to be the primary contributor to membership inference accuracy. Our analysis takes a step further by examining the relation between MIA and several ML-model design attributes, and by exploring the dependence of MIA on model characteristics other than overfitting. The model properties considered in this study are described below.

\textit{Model Architecture and Training Configuration.} ~Using an Artificial Neural Network (ANN) as our default classifier type, we explore two main architectural properties: the model \textit{depth} (the number of hidden layers) and the size of each layer (number of neurons). We also include in our analysis a few key training parameters, such as the learning rate and the learning regularizer.    

\textit{Target-Shadow Model Combination.} ~Following an approach similar to ~\cite{18truex2018towards}, we consider various widely-used classifier types to understand the effect of different target-shadow model combinations on MIA: Logistic Regression (LR), Support Vector Machine (SVM), Random Forest (RF), $k$-Nearest Neighbour ($k$NN) and ANN.

\textit{Mutual Information between model parameters and training records.} ~This property quantifies the information extracted by the target model from the set of training records. Assume a generic ML model $f(X;\theta)$, whose training is conducted over records $x_i \in X$ and generates the model's parameters $\theta$.
For such a model, the mutual information $I_X$ is obtained as:
\begin{equation}\label{eq:mi}
    I_X = \frac{1}{|X|} \sum_{x_i\in X} I(x_i;\theta),
\end{equation}
where $I(x_i;\theta)$ is the mutual information between an individual training record and $\theta$, computed as below:
\begin{equation}
    I(x_i;\theta)= H(x_i)-H(x_i|\theta),
\end{equation}
where $H(x_i)$ is the entropy of one record and $H(x_i|\theta)$ is the conditional entropy that captures the uncertainty in $x_i$ when the parameter values $\theta$ are known.

\textit{Deviations from perfect MIA-indistinguishability.} ~The perfect MIA-indistinguishability property defined in~\cite{yaghini2019disparate} is a guarantee of perfect model resistance against MIA. It indicates that, for a generic prediction output $y'$, the input record has the same probability to be a member or a non-member of the training dataset: 
  \begin{equation}\label{eq:mem_non}
    P[m=1| \hat{y} =y'] = P[ m=0| \hat{y}=y'],
    \end{equation}
where $m \in \{0,1\}$ is the membership value denoting whether the record is a member ($m=1$) or non-member ($m=0$) of the training set. Our analysis explores how increasing deviations from perfect MIA-indistinguishability affect the ML model resilience to MIA. We measure such deviations as the $l_\infty$-relative metric between the considered probabilities and maximum divergence across the classes of $y$ \cite{yaghini2019disparate}:
\begin{equation}\label{eq:membership_measure}
    \delta_{mi}(f) :=\sup_{y'\in y}  |\log \{\frac{P[\hat{y}=y'|m=1]}{P[\hat{y}=y'|m=0]}\}|
\end{equation}
A model is presumably less vulnerable to MIA if $\delta_{mi}$ is close to zero.

\textit{Model Fairness.} ~We consider the model's prediction fairness towards different individuals or subgroups of the datasets based on a set of particular protected features. For examples, assume `Gender' is a protected feature in a dataset containing two feature values: `male' and `female'. The fairness with respect to this protected features indicates how \textit{equally} the model treats `male' and `female' records in terms of prediction output.
Among the existing definitions of ML fairness~\cite{gajane2017formalizing, verma2018fairness, binns2017fairness, barocas2017fairness}, we consider the (i) group (or statistical) fairness, (ii) predictive fairness and (iii) individual fairness. 

\begin{itemize}

    \item \textit{Group fairness.} ~\gio{A classifier $f(X)\implies y$ is fair towards two groups of records $g_i, g_j \subset X$ if it predicts a particular outcome for the individuals across the protected subgroups with 
    equal probabilities~\cite{gajane2017formalizing, verma2018fairness}:
      \begin{equation} \label{eq:group}
          P[\hat{y}_{i} = y' | x_i \in g_i, y'\in y] =  P[\hat{y}_{j} = y'| x_j \in g_j, y'\in y],
      \end{equation}
    where $\hat{y}_{i}$ and $\hat{y}_{j}$ are the predicted outcomes for the records in groups $g_i$ and $g_j$ respectively and 
    $y'\in y$ is the preferred class label that needs to be predicted fairly for all groups of records. This definition is also known as demographic parity. To quantify the group fairness, we estimate the probabilistic difference of prediction $\delta_g$ between two subgroups of the records for $n$ number of class labels as below:
        \begin{equation} \label{eq:group_measure}
            \delta_g(f) := \frac{1}{n}\sum_{y'\in y} |P[\hat{y}_{i}=y'] - P[\hat{y}_{j}=y']|
        \end{equation}
    }
   
    \item{\textit{Predictive fairness.} \gio{~A classifier has a predictive fairness if the subgroups of the records that truly belong to the 
    preferred class have equal probabilities to be predicted in that 
    class \cite{verma2018fairness}. 
    That is,}
      
      {\small
      \begin{equation}\label{eq:pred}
          P[\hat{y_i}=y' | y_i=y', x_i \in g_i] =  P[\hat{y_j}=y' | y_j=y', x_j \in g_j],
      \end{equation}
      }
      
    where $y'$  is the preferred class label. This is also known as Equal Opportunity in the literature. For $n$ class labels, the deviation from perfect predictive fairness $\delta_p$ can be measured as:
      
      \begin{equation} \label{eq:pred_measure}
            \delta_p(f) := \frac{1}{n} \sum_{y' \in y} |P[\hat{y_i}=y'|y_i=y']- P[\hat{y_j}=y'|y_j=y']|
        \end{equation}
    }
    
    \item{\textit{Individual fairness.} \gio{~This concept, introduced in \cite{dwork2012fairness}, indicates that the predictor produces similar prediction outputs for similar inputs. For a ML classifier, the condition of individual fairness can be formalized as follows~\cite{gajane2017formalizing}:
    \begin{equation} \label{eq:ind_fair}
        \Delta(\hat{y}_{x_i}, \hat{y}_{x_j}) \leq d(x_i,x_j)
    \end{equation}
    where, $d(x_i,x_j)$ is the distance between two generic input records $x_i$ and $x_j$, and $\Delta(\hat{y}_{x_i}, \hat{y}_{x_j})$ is the distance between the corresponding prediction outputs.} 
    \gio{For a given dataset $D(X,y)$, the overall} $\Delta$ and $d$ can be measured in different ways. In \cite{dwork2012fairness}, a statistical distance metric is proposed for $\Delta$ that measures the Total Variation (TV) norm distance between two probabilities for the outcome of the classifier for the two considered records:  
        \begin{equation} \label{eq:dwrok_ind}
            \Delta_{tv}=\frac{1}{n} \sum_{\hat{y}_{x_i}, \hat{y}_{x_j} \in y} |P[\hat{y}_{x_i}]-P[\hat{y}_{x_j}]| 
        \end{equation}
    This metric assumes that the distance metric selected for $d$ will scale the measured distance within $[0,1]$ range. Authors of \cite{dwork2012fairness} also suggest that a better choice for $\Delta$ is to use the relative $l_\infty$ norm metric
which would allow the use of a metric for $d$ that considers two records to be similar if $d<<1$ and dissimilar if $d>>1$. To keep both the $d$ and $\Delta$ within the same range [0,1], 
we measure their statistical distances in both the cases.  
Finally, we compute the deviation from individual fairness $\delta_i$ as follows:
    \begin{equation} \label{eq:ind_measure}
 	     \delta_i(f) := \sum_{x_i, x_j \in X}|\Delta_{tv}- |P[x_i]-P[x_j]||
    \end{equation}
    }
\end{itemize}

\subsection{Experimental Setup}

Next we present the datasets in use and the ML models for MIA. 



\textbf{Datasets.}
~We consider three datasets widely used for the analysis of MIA~\cite{10shokri2017membership,truex2019effects, 18truex2018towards, attriguard,nasr2018machine, schonherr2018adversarial}: Purchase, Texas hospital, and UCI Adult datasets.



Primarily, our datasets are pre-processed as below:


\begin{enumerate}
    \item Purchase dataset~\cite{kaggle}. The Purchase dataset includes two subsets: the customer transactions and the history of the incentives offered to them. To have a complete dataset, we join the two subsets by the customer IDs. 
    After the join operation, we have a dataset with 16 features. We prepare the primary dataset by uniformly sampling $400,000$ records.
    
    
    
    \item Texas hospital dataset~\cite{texas}. 
    \shakila{We avoid the features with missing values and pre-process this dataset by selecting 16 features} \zhigang{(same number of features as the Purchase dataset)}. 
    \zhigang
    {For the primary dataset, we also randomly sample $400,000$ records
    from years 2006 to 2009}.
    
    
    
    \item UCI Adult dataset~\cite{uci_adult}\zhigang{. This is a small dataset containing} $48,842$ \zhigang{records} with $14$ features. \zhigang{We use it as the primary dataset directly in our experiments.} 
    
\end{enumerate}

For all the datasets, we restrict the set of output classes to two (label 0 and 1). By focusing on binary datasets, we simplify the analysis of data properties such as the  balance in classes. Further more, we do not consider image datasets (e.g., CIFAR-10~\cite{cifar_2009}) due to the difficulties these datasets would pose on the analysis of feature balance. In particular, the number of features would be very high when considering the raw image pixels, while the adoption of more elaborate features (e.g., edges, shapes, motion) would make it difficult to explore different feature-balance ranges. Overall, we choose three datasets that have been extensively used in previous work, and which are suitable to exploring a wide range of properties.

\zhigang{\textbf{ML Models.}} 
~By default, we implement the target, shadow, and attack models using Artificial Neural Networks (ANN) with the hyperparameters setup reported in Table~\ref{tab:model_parameters}. We fix the \textit{member rate} to 75\%, \textit{i.e.}, for every experiment the 75\% of dataset records are members of the training set. Our default ML model achieves 85\% train 84\% test accuracy on Texas dataset, (76\%, 75\%) accuracy on Purchase, and (70\%, 71\%) accuracy on Adult.
To reduce the computational complexity and thereby extend the scale of our experiments, we perform the attack using by default only one shadow model: this approach builds on the findings in \cite{salem2018ml}, i.e., that similar MIA accuracy is achieved using one instead of multiple shadow models. All ML models are trained and tested using Theano~\cite{theano_2008} and Lasagne~\cite{lasagne_2015} libraries.



\begin{table}[ht!]
    \centering
    \captionsetup{justification=centering}
    \subfloat[Default model setup.]{
    \centering
    \begin{tabular*}{.47\textwidth}{ll}
    Model & Hyperparameters\\
    \hline
    Artificial Neural Network (ANN) & $\alpha$ = 0.001, solver= sgd,\\
    & epochs=50,$\lambda$=1e-7 \\
    & 1-hidden layer, 50 neurons\\
    \end{tabular*}
    }
    \hfill
    \subfloat[Additional model setups.]{
    \centering
    \begin{tabular*}{.47\textwidth}{ll}
    Models & Hyperparameters\\
    \hline
    Logistic Regression (LR) & $C$=0.01, solver= LBGFS \\
    Support Vector Machine (SVM) & $C$=0.01, kernel= RBF \\
    Random Forest (RF) & n-estimators=100, \\ 
    & criterion=gini, max-depth=2 \\
    $k$-Nearest Neighbour ($k$NN) & $p$=2, neighbors=3 \\
    \end{tabular*}
    }
    \caption{Selection of the hyperparameters for the ML models.}
\label{tab:model_parameters}
\end{table}

\section{Analysis and Results}
\label{sec:result}
This section experimentally explores the relation between MIA and the different properties described in Section~\ref{sec:data}. To evaluate MIA's performance, we measure the accuracy of the attack model (\textit{MIA accuracy}), which corresponds to the fraction of correct membership predictions. To show the uncertainty on MIA accuracy results, we provide the average MIA accuracy along with the 95\% confidence interval (error bars). 

\subsection{Effect of Data Properties on MIA}\label{sec:data_prop}
\textbf{The Size of Dataset:}~In our experiments, we report the average results of the MIA accuracy when varying the size of the target dataset and the shadow dataset respectively. 
Particularly, to tune the size of dataset (either the target dataset or the shadow dataset), we sample subsets from the primary datasets in the range [10K,100K] for both Purchase and Texas datasets and [1K,10K] for Adult dataset.

\begin{figure*}[t]
    \centering
    \captionsetup{justification=centering}
    \subfloat[Purchase dataset.]{
        \includegraphics[width=.29\textwidth]{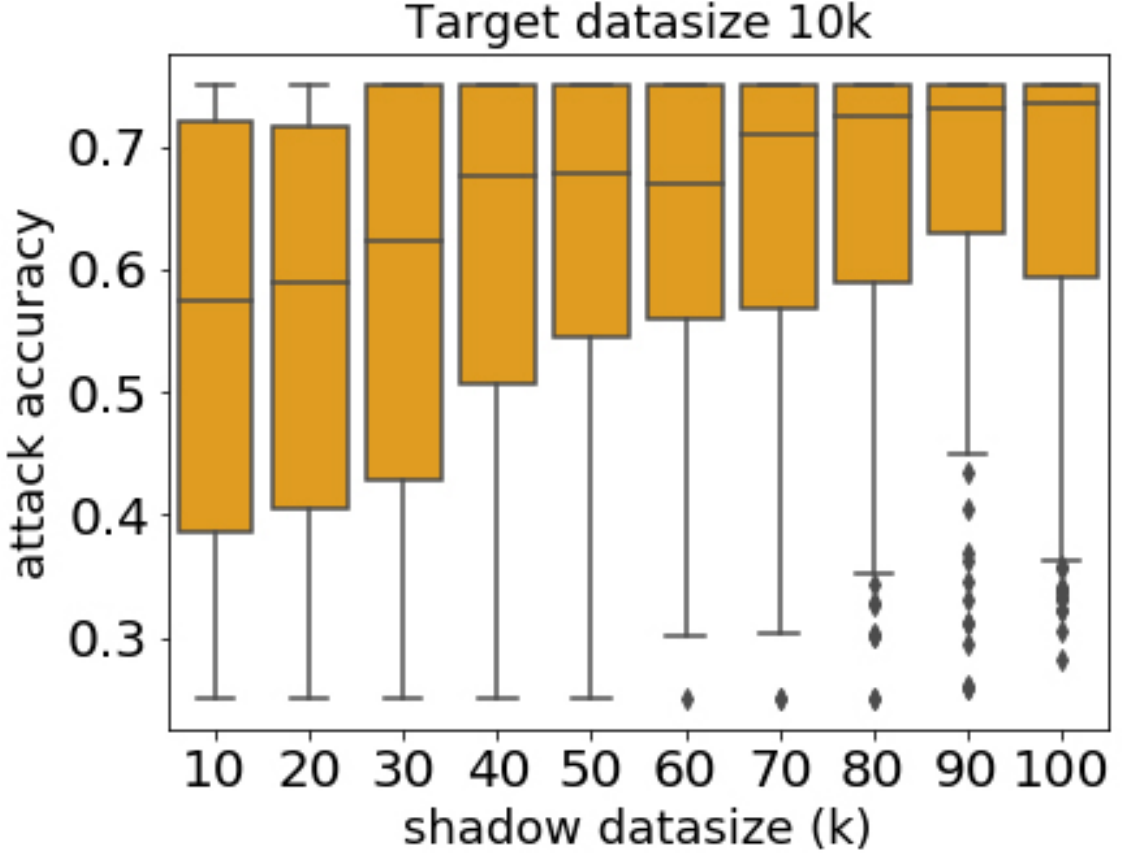}
    }
    \hfill
    \subfloat[Texas dataset.]{
        \includegraphics[width=.29\textwidth]{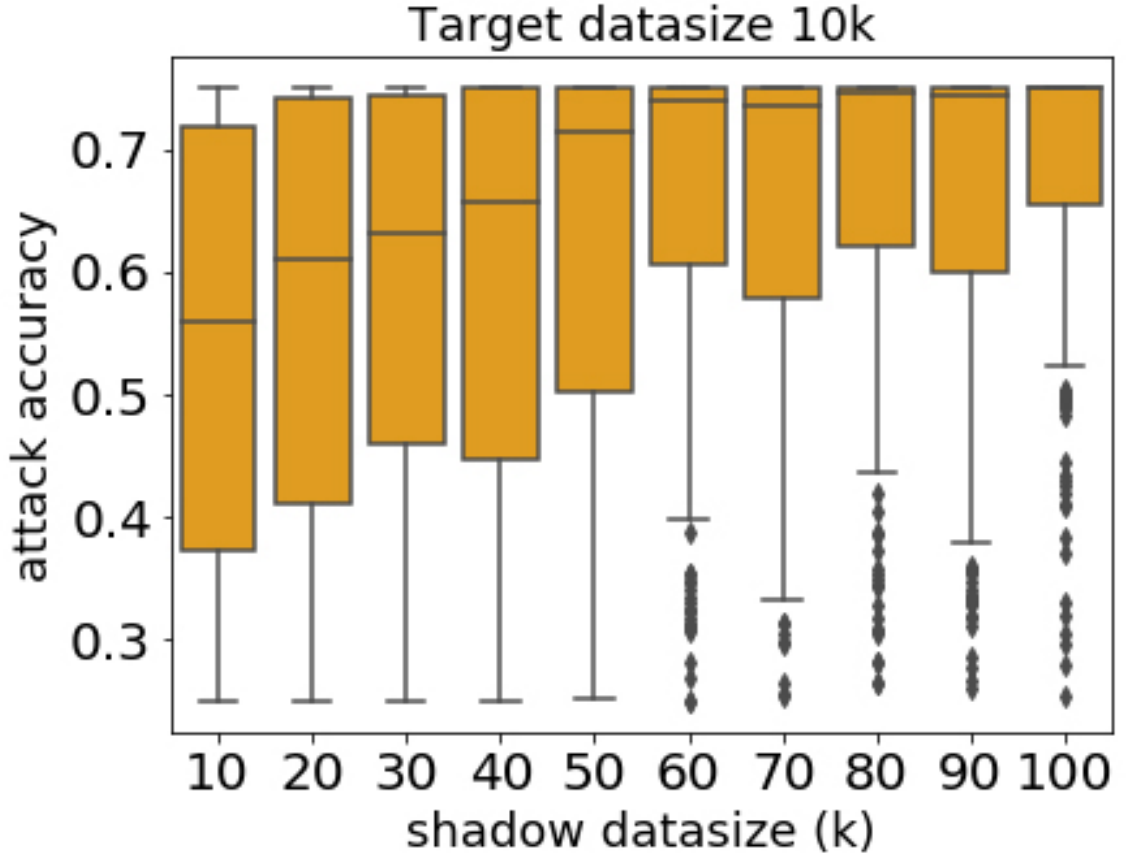}
    }
    \hfill
    \subfloat[Adult dataset.]{
        \includegraphics[width=.29\textwidth]{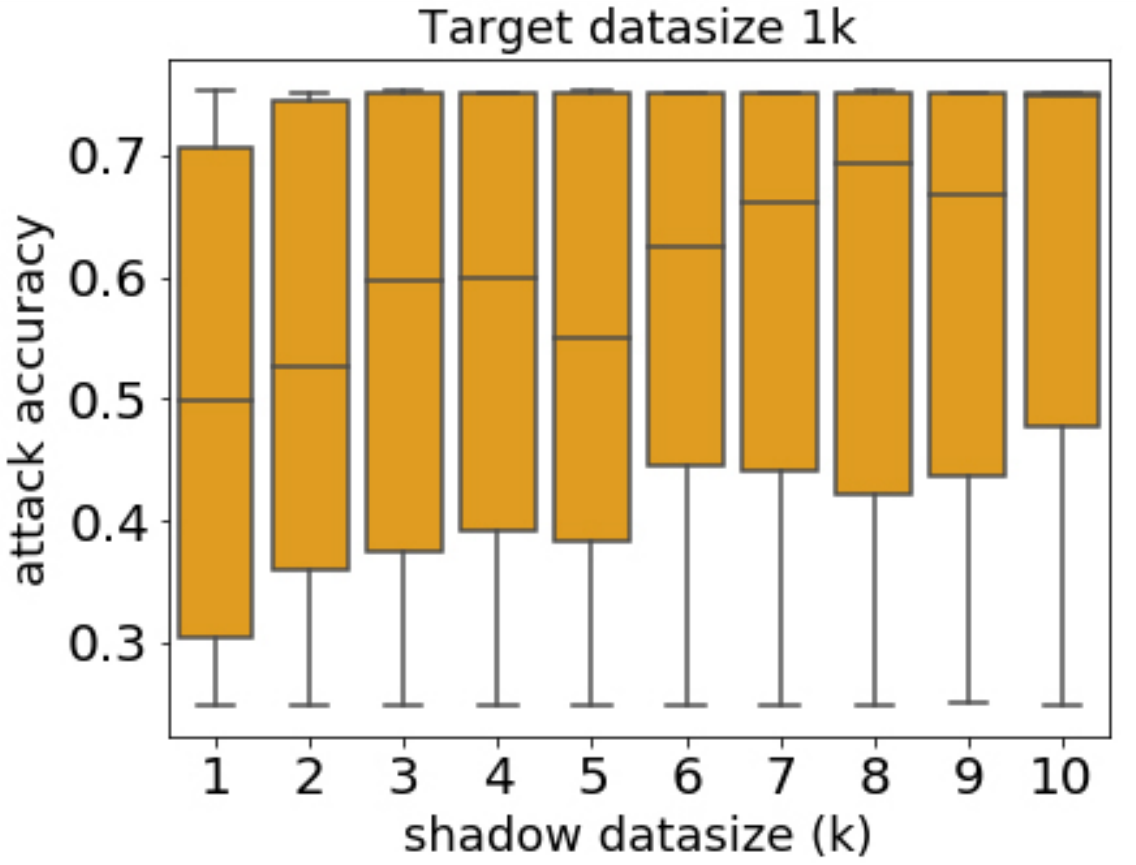}
    }
    \caption{Effects of varying the size of the shadow datasets on MIA accuracy.}
    \label{fig:datasize_1}
\end{figure*}

\begin{figure*}[t]
    \centering
    \captionsetup{justification=centering}
    \subfloat[Purchase dataset.]{
        \includegraphics[width=.29\textwidth]{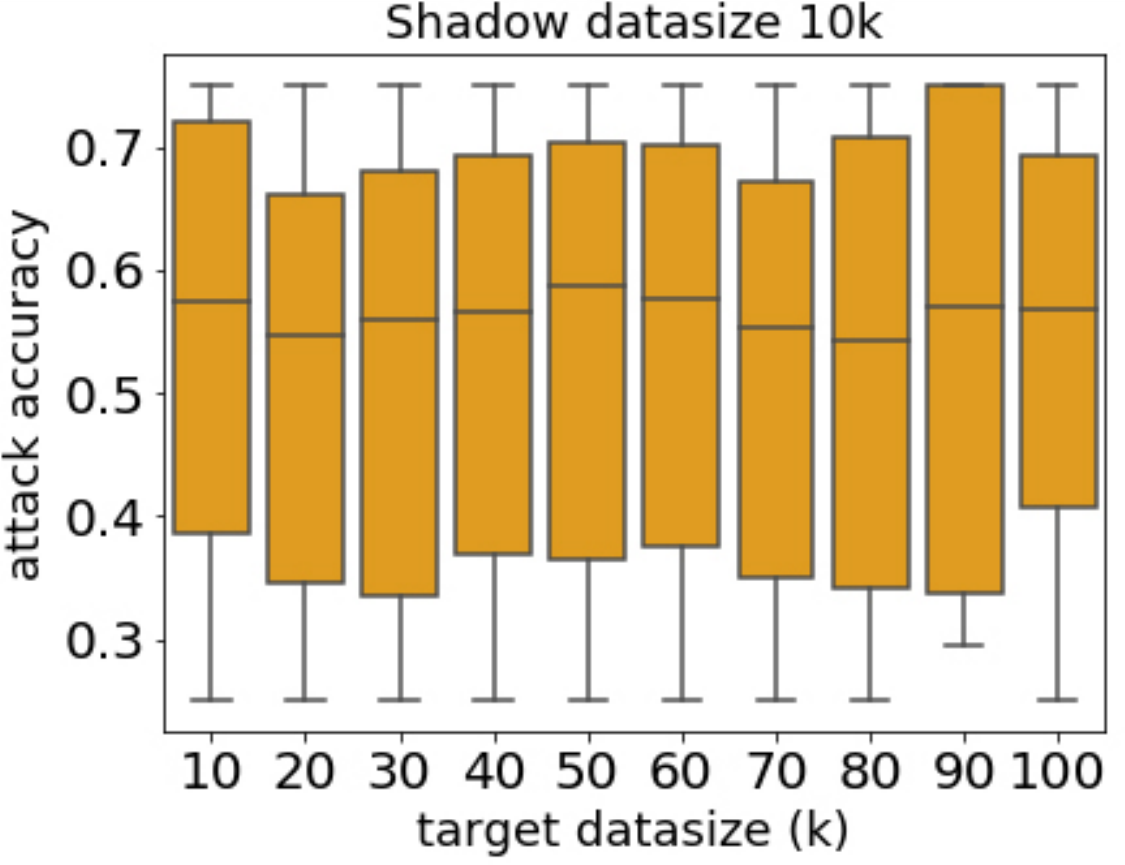}
    }
    \hfill
    \subfloat[Texas dataset.]{
        \includegraphics[width=.29\textwidth]{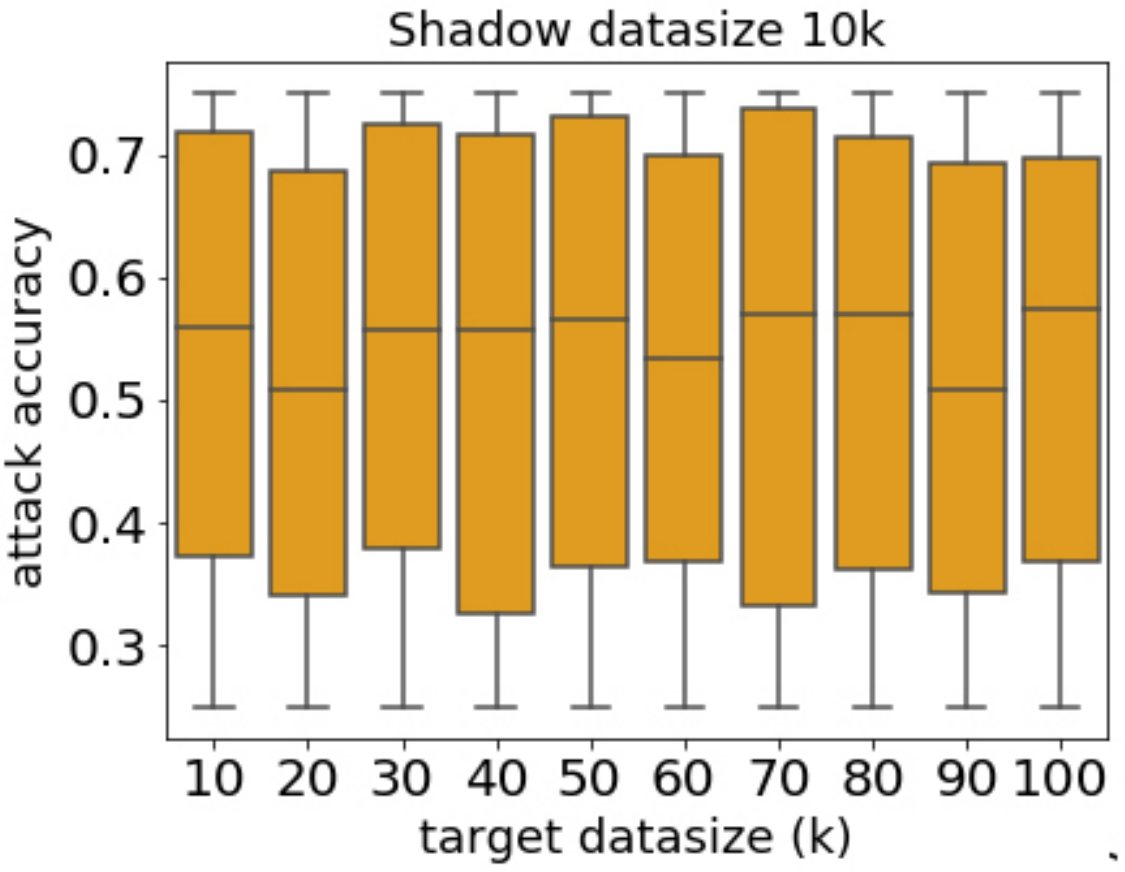}
    }
    \hfill
    \subfloat[Adult dataset.]{
        \includegraphics[width=.29\textwidth]{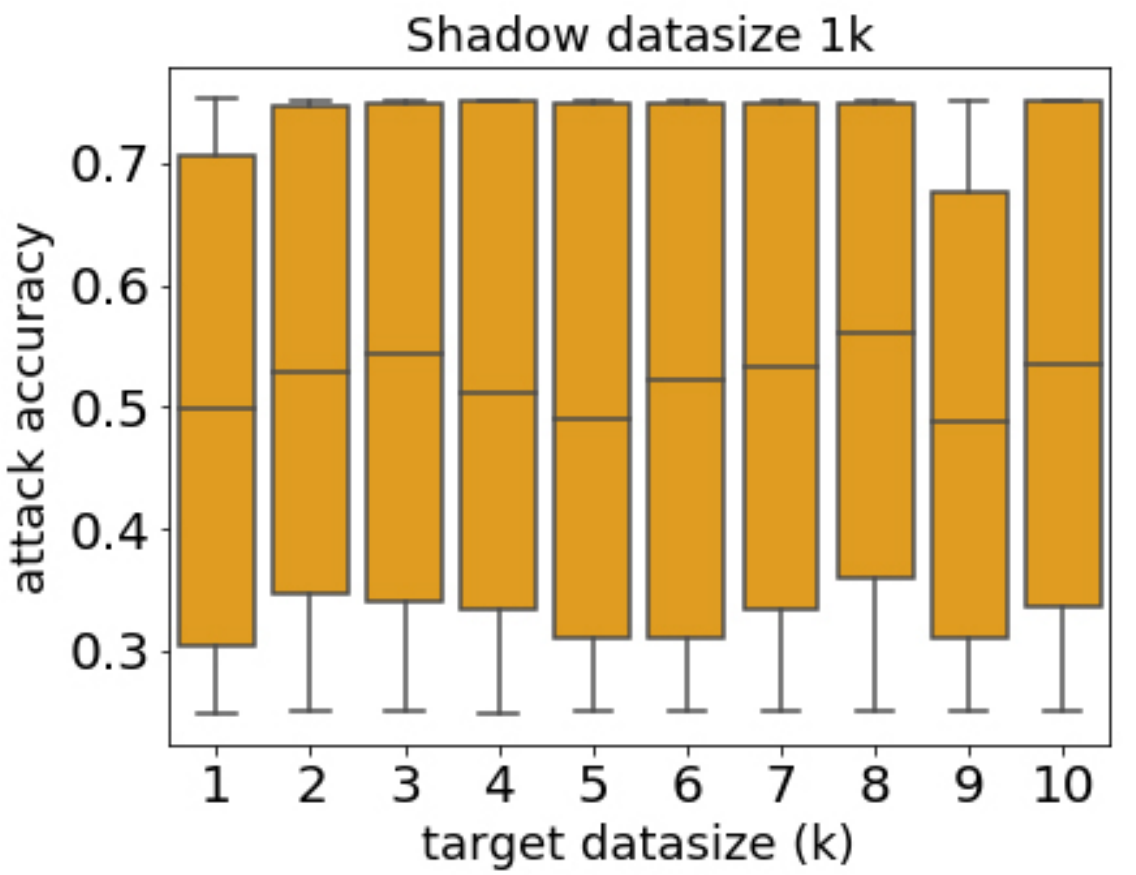}
    }
    \caption{Effects of varying the size of the target datasets on MIA accuracy.}
    \label{fig:datasize_2}
\end{figure*}



First, from Figure~\ref{fig:datasize_1}, we observe that with a fixed target dataset size, the size of shadow datasets has a positive impact on MIA accuracy. That is, by increasing the size of Shadow dataset, the average MIA accuracy increases from about 0.5 to 1 over all the three datasets.  


In contrast, as illustrated in Figure~\ref{fig:datasize_2}, when fixing the size of the shadow datasets, the size of target dataset does not show any significant impact on MIA accuracy. This suggest that the size of shadow datasets indicates and bounds the advantage/power of MIA adversaries against a given target dataset.

This result confirms how overfitting impacts the MIA accuracy from a new perspective. It is known that a relatively small target training dataset produces higher overfitted models~\cite{yeom2018privacy}. 
Our results therefore validate the findings of some of the existing works which have shown that overfitting is a major reason to the success of MIA, as the size of shadow models (and hence the overfitting) has a positive effect on MIA performance.

\textbf{The Balance in Classes: } 
In our experiments, we convert all datasets to binary-class datasets. This way, we can measure the impact of class balance by tuning just one class balance value. That is, a class balance value x\% indicates a x/(100-x) or (100-x)/x split between classes 0 and 1, with 50\% representing a perfect balance. Specifically, we assign two classes (0,1) to the records of the Purchase and Texas datasets by using Gaussian mixture clustering~\cite{reynolds2009gaussian, gauss_2007}  
and the Adult dataset already contains two classes based on the salary feature.

\begin{figure*}[ht!]
    \centering
    \captionsetup{justification=centering}
    \subfloat[Purchase dataset.]{
        \includegraphics[width=.29\textwidth]{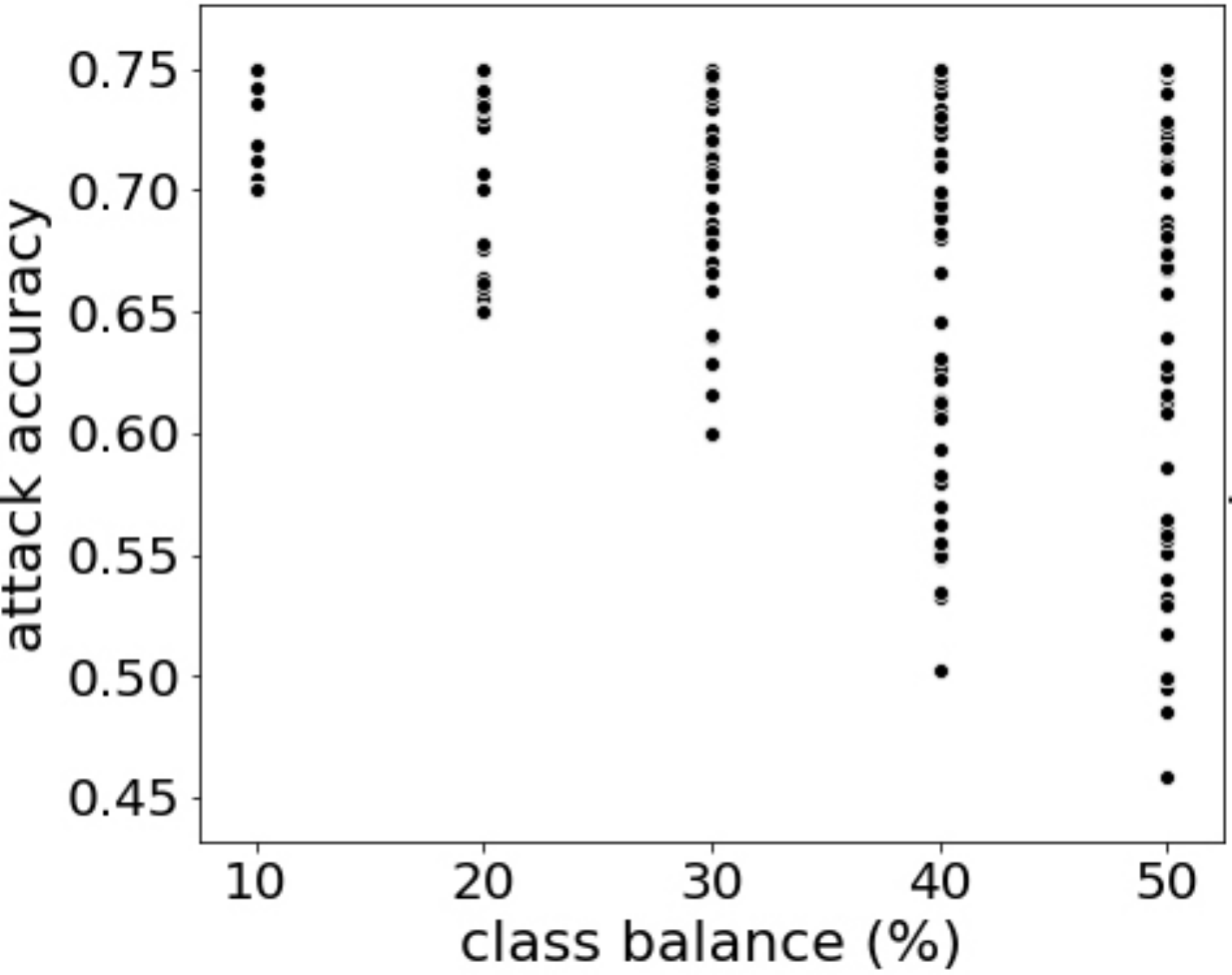}
    }
    \hfill
    \subfloat[Texas dataset.]{
        \includegraphics[width=.29\textwidth]{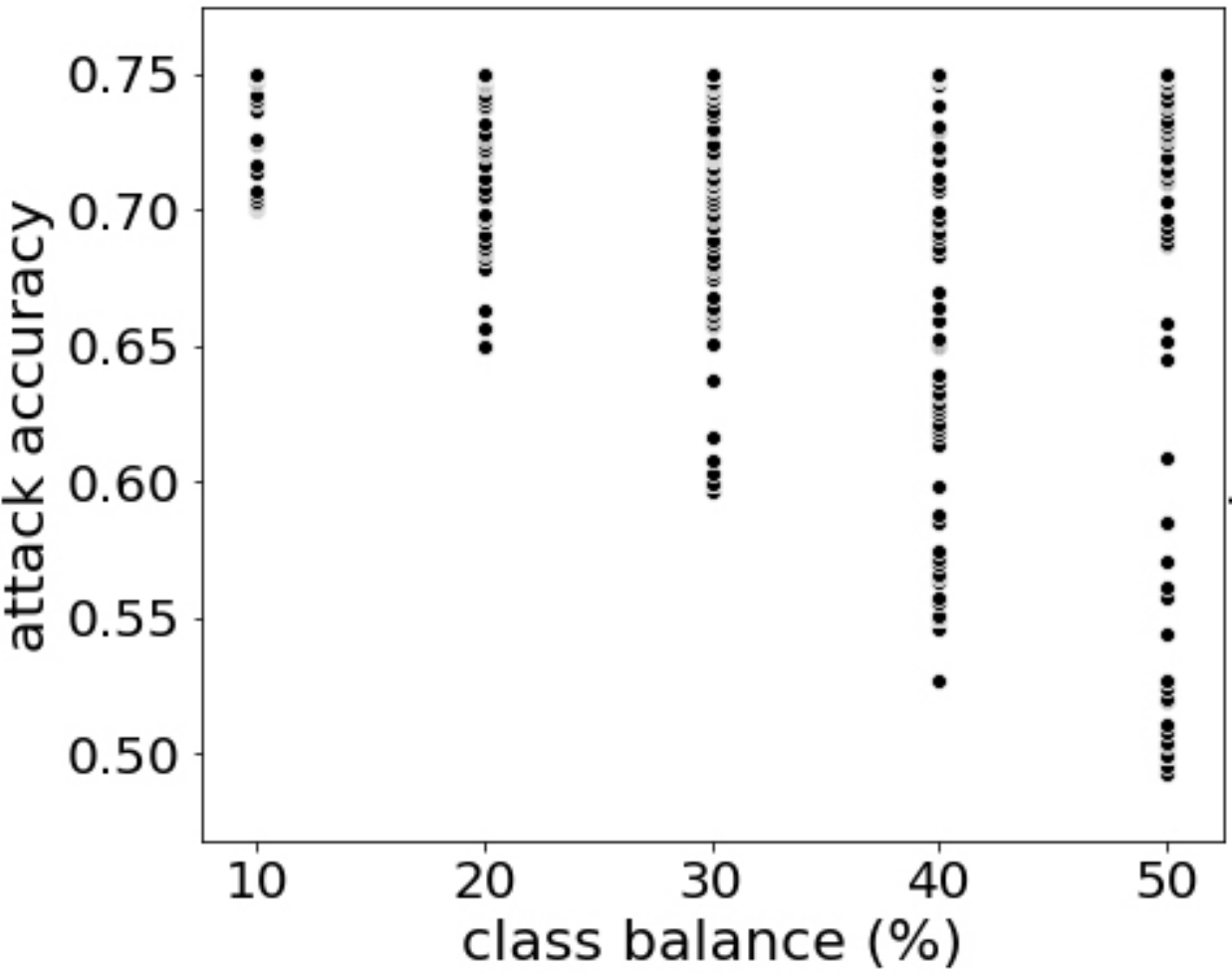}
    }
    \hfill
    \subfloat[Adult dataset.]{
        \includegraphics[width=.29\textwidth]{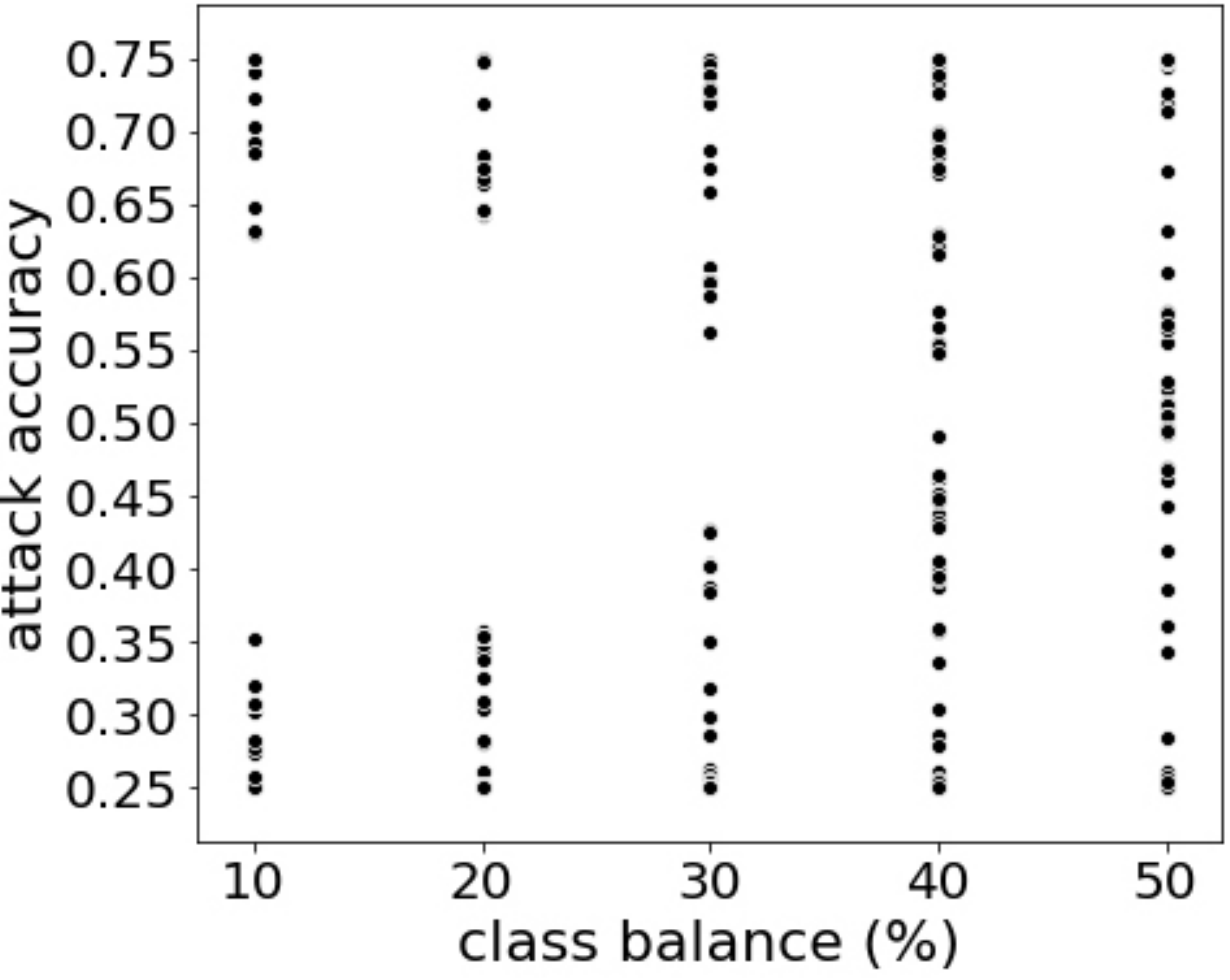}
    }
    \caption{Effects of the balance in the classes on MIA accuracy. 
    }
    \label{fig:p_class_attack}
\end{figure*}

Figure~\ref{fig:p_class_attack} depicts the effect of class balance on the MIA performance over the three datasets. In the sub-figures, each dot shows the MIA performance of one experiment from a target training dataset with a given class balance. The figure shows the results for 100 experiments for each class balance, and as such represents the distribution of MIA accuracy across different class balance values.

 


From Figure~\ref{fig:p_class_attack}, we discover that with an increasing class balance, the minimum value of MIA accuracy is decreasing from about 0.7 to 0.5 in all three datasets. Such results reflect the truth that the class balance measures (and has a negative impacts on) the lower bound of MIA accuracy.
We also note that similar results are also reported in \cite{truex2019effects}, where authors observed that the skewness of the training dataset plays a role in increasing MIA performance, as it becomes difficult for the model to generalize the prediction over a class of fewer number of instances. This behaviour can be explained by the target model's bias in prediction. The model produces less biased prediction towards one particular class when it is trained on a dataset that has properly balanced classes giving less advantage to an adversary. 



\begin{figure*}[ht!]
    \centering
    \captionsetup{justification=centering}
    \subfloat[Purchase dataset.]{
        \includegraphics[width=.29\textwidth]{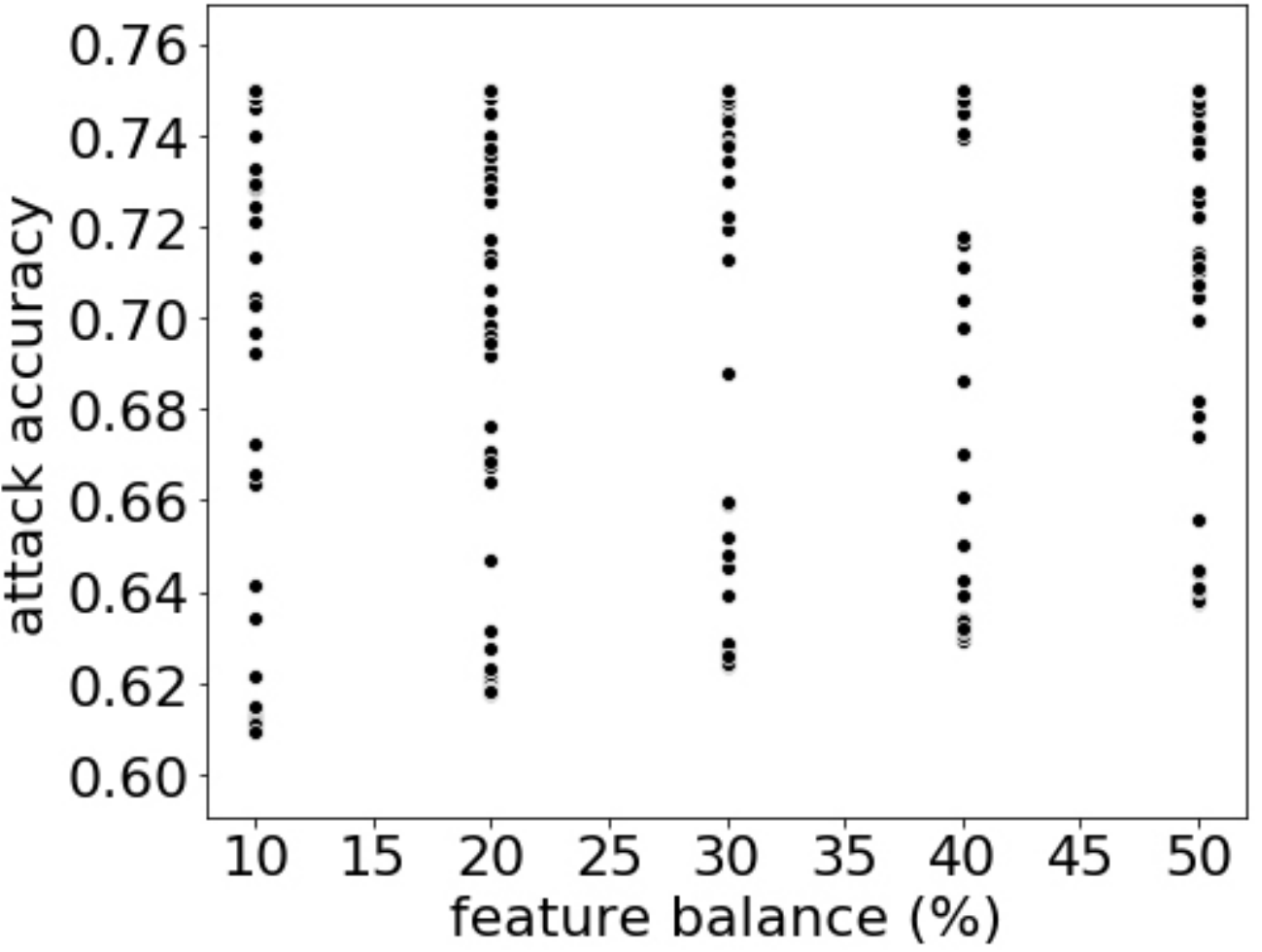}
    }
    \hfill
    \subfloat[Texas dataset.]{
        \includegraphics[width=.29\textwidth]{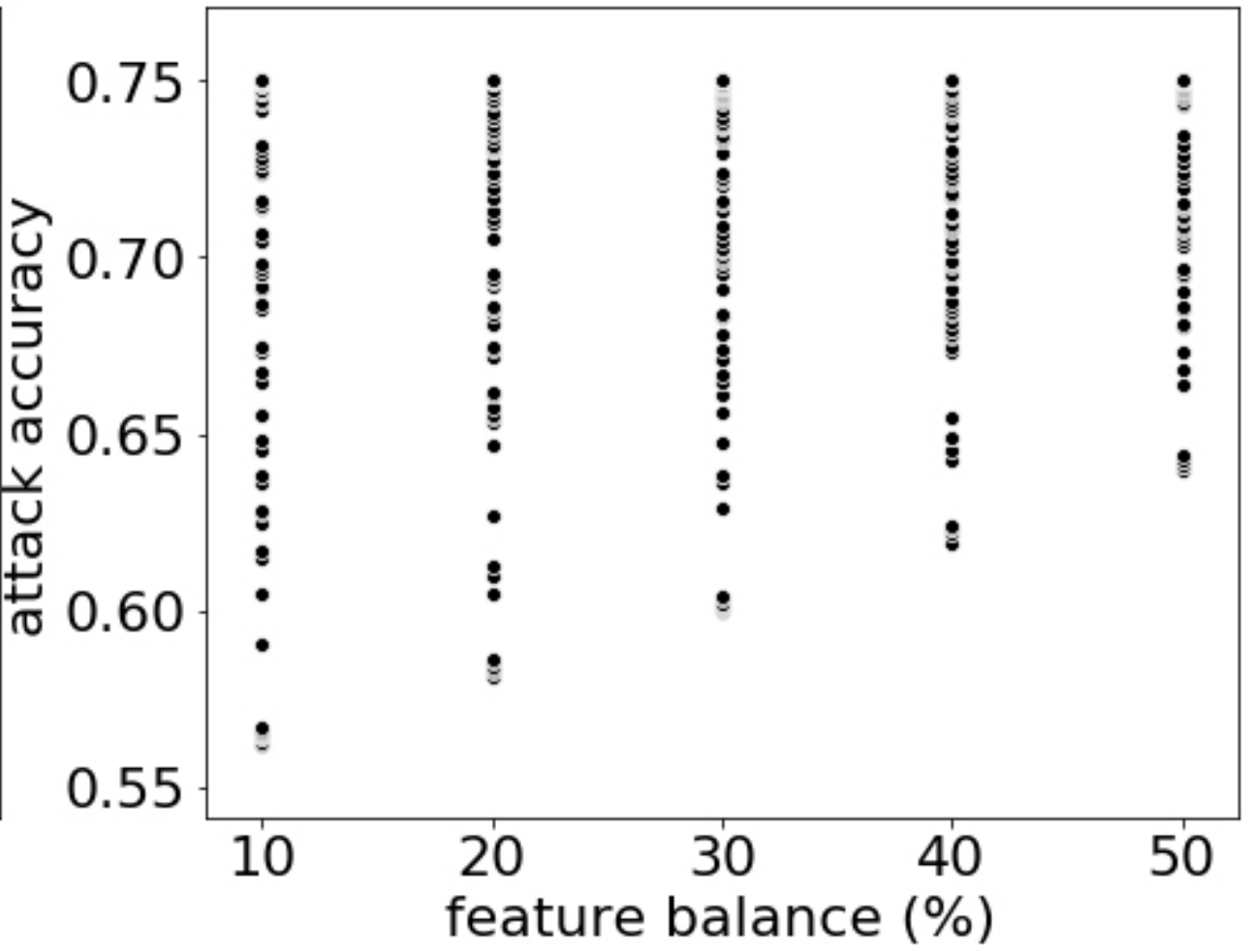}
    }
    \hfill
    \subfloat[Adult dataset.]{
        \includegraphics[width=.29\textwidth]{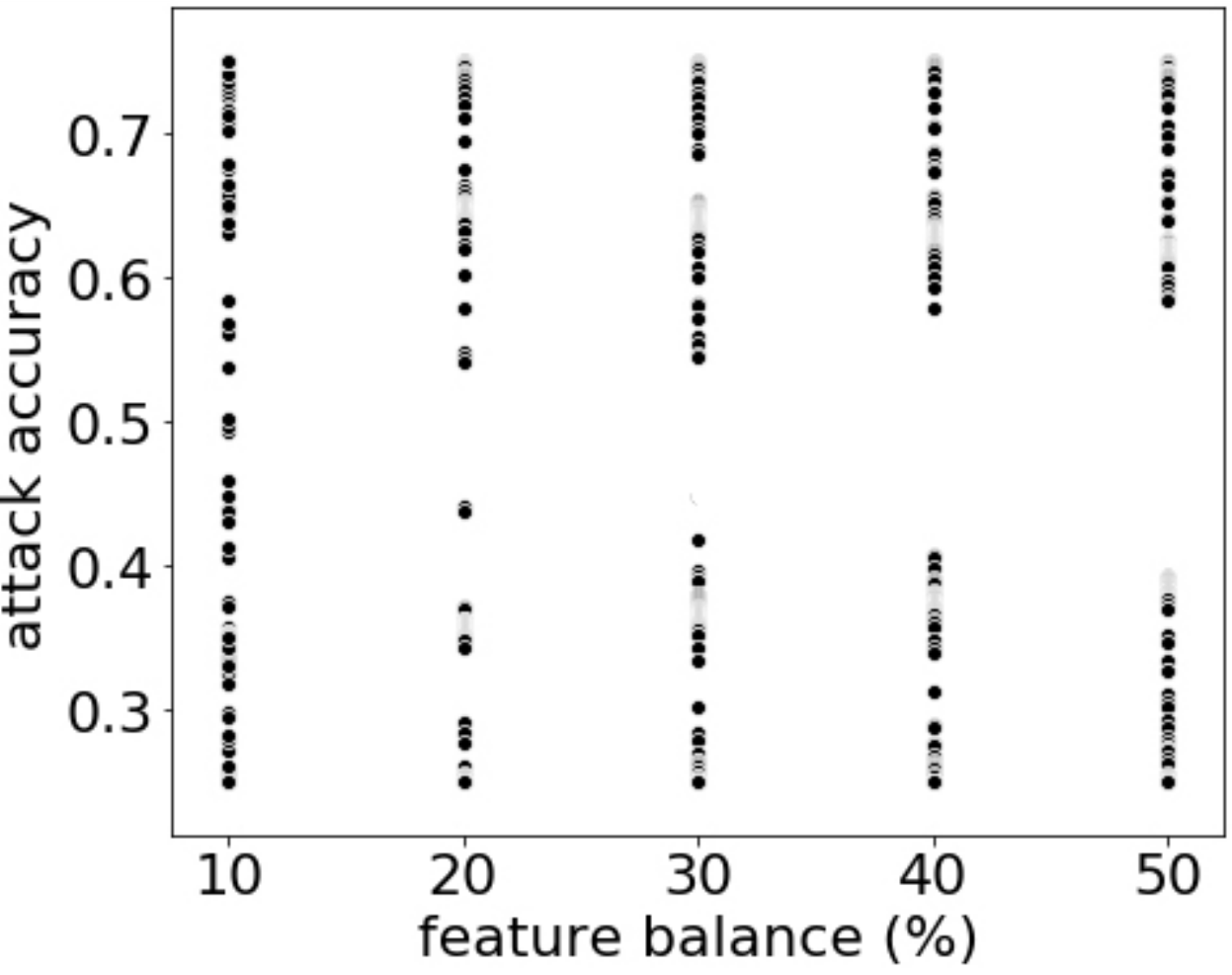}
    }
    \caption{\zhigang{Effects of the balance in the features on MIA accuracy.}
    }
    \label{fig:p_feature_attack}
\end{figure*}

\textbf{The Balance in Features: }
For each of the datasets, we select the 5 features that have the highest 
mutual information with the class labels and measure the feature balances as the percentage probability ratio of the selected combination of the feature values.

Figure \ref{fig:p_feature_attack} shows the performance of the attack on the datasets 
for the class balances $<50\%$ with the selected 5 features. 
Similarly to our experiments in Figure~\ref{fig:p_class_attack}, we run 100 experiments for each feature balance so that we report the distribution of MIA performance over different values of feature balance. The results show an increase in the lower bound of MIA accuracy for an increased imbalance of the features. 


An increase of the features balance results in increasing the number of data points that have similar feature values. As a result, the target model's prediction becomes more biased towards dominant feature values and as such correct guesses from an MIA attacker, given a target, become  more likely.

\begin{figure*}[ht!]
    \centering
    \captionsetup{justification=centering}
    \subfloat[Purchase dataset.]{
        \includegraphics[width=.29\textwidth]{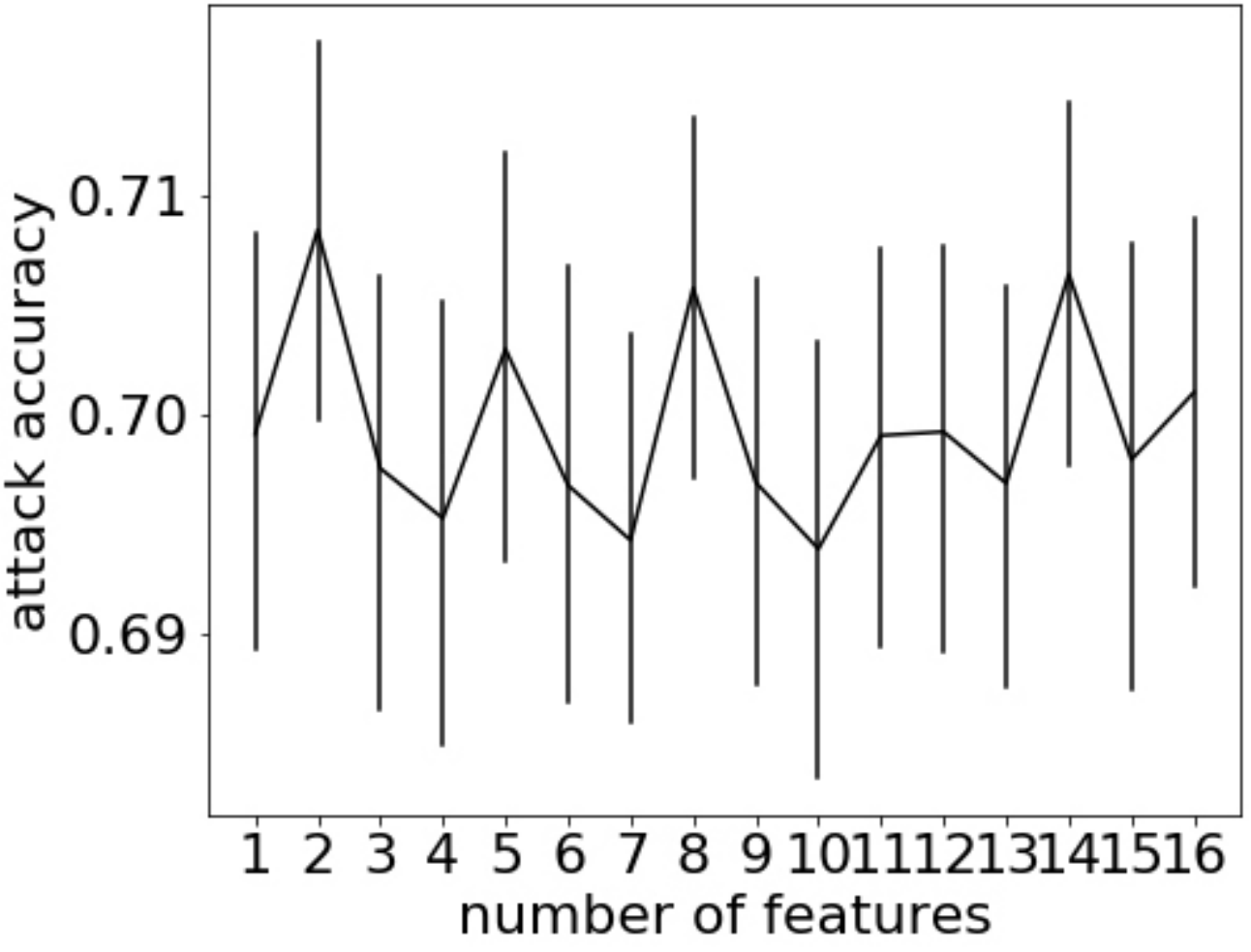}
    }
    \hfill
    \subfloat[Texas dataset.]{
        \includegraphics[width=.29\textwidth]{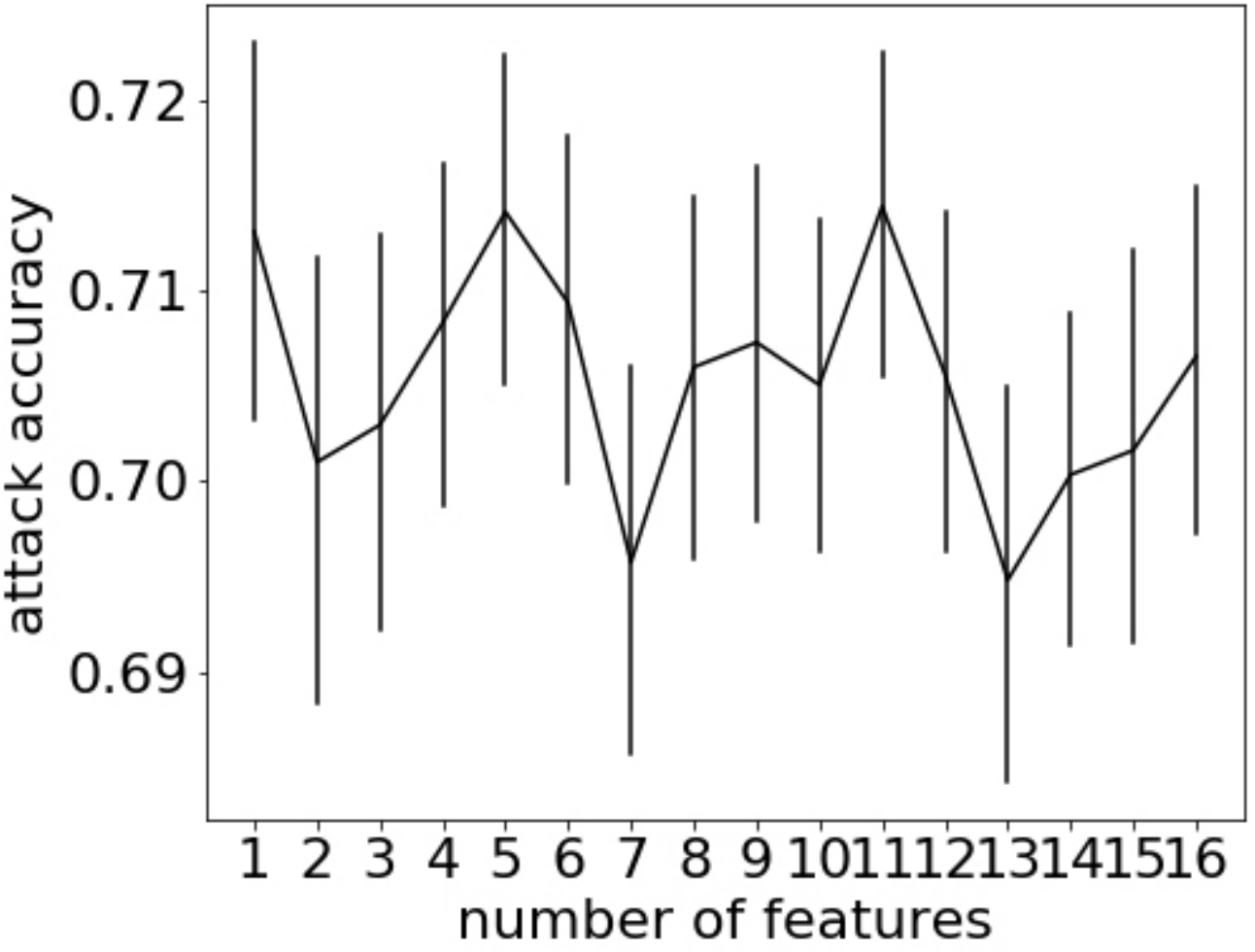}
    }
    \hfill
    \subfloat[Adult dataset.]{
        \includegraphics[width=.29\textwidth]{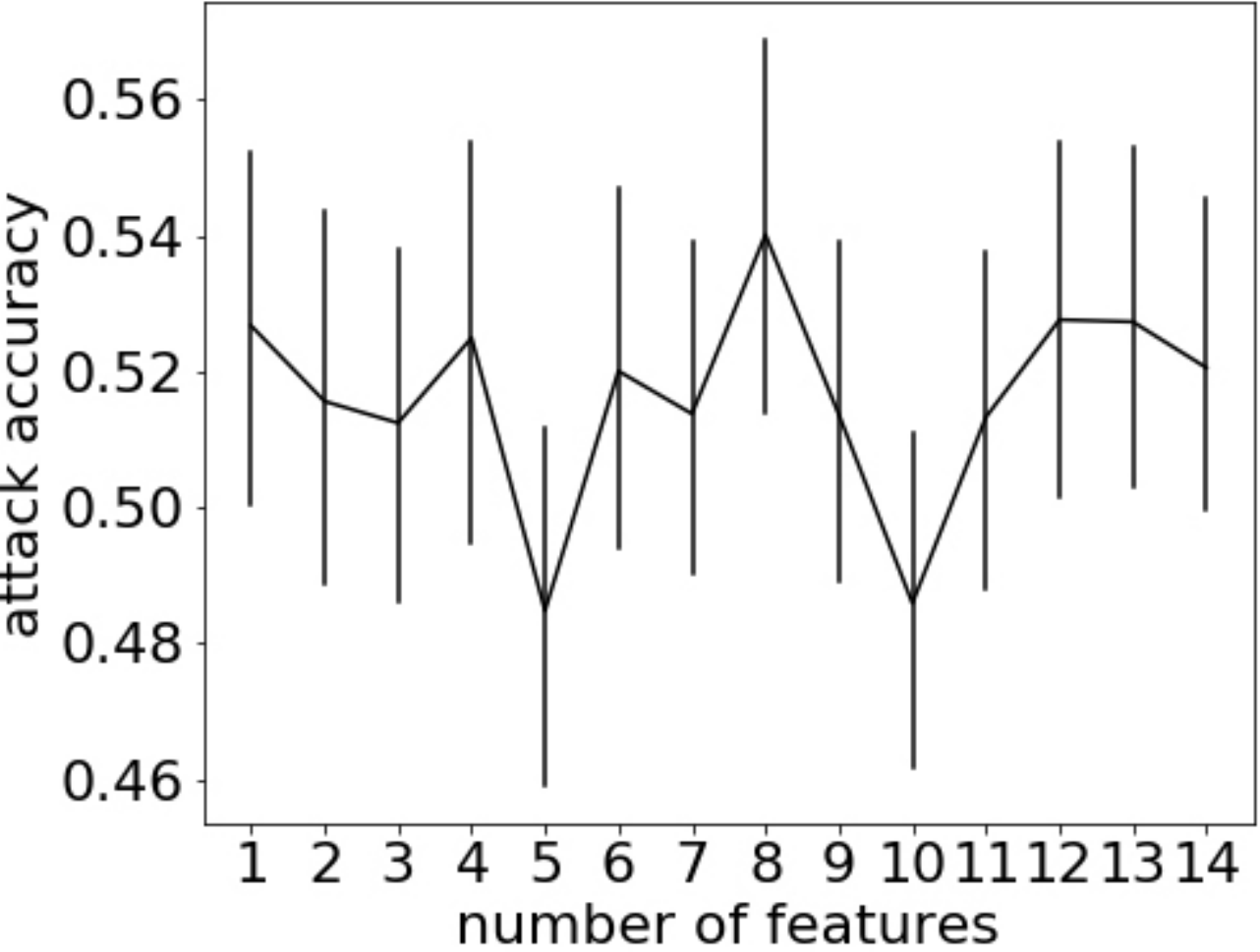}
    }
    \caption{\zhigang{Effects of the number of features on MIA accuracy.} 
    }
    \label{fig:feature_no}
\end{figure*}

\textbf{The Number of Features:} 
~Figure \ref{fig:feature_no} represents the observed attack accuracy for the datasets by gradually increasing  the number of features. In our experiments, we vary the number of features while the specific features in each experiment are randomly selected. 

The figure suggests the number of features of a training dataset does not have any effect on the MIA accuracy. This indicates that we could leverage the  number of features to achieve higher ML utility without sacrificing the privacy disclosure against MIA.


\begin{figure*}[ht!]
    \centering
    \captionsetup{justification=centering}
    \subfloat[Purchase dataset.]{
        \includegraphics[width=.29\textwidth]{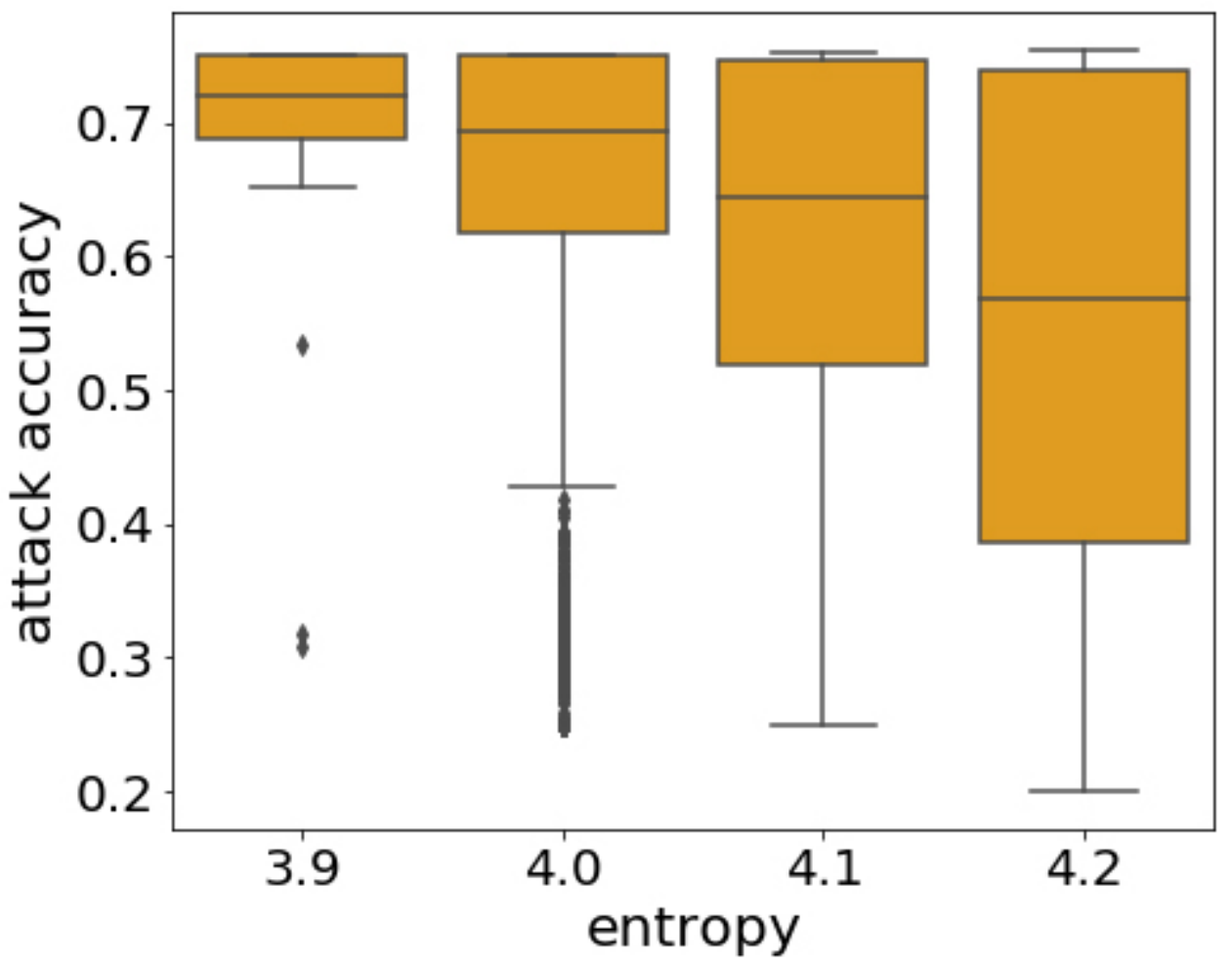}
    }
    \hfill
    \subfloat[Texas dataset.]{
        \includegraphics[width=.29\textwidth]{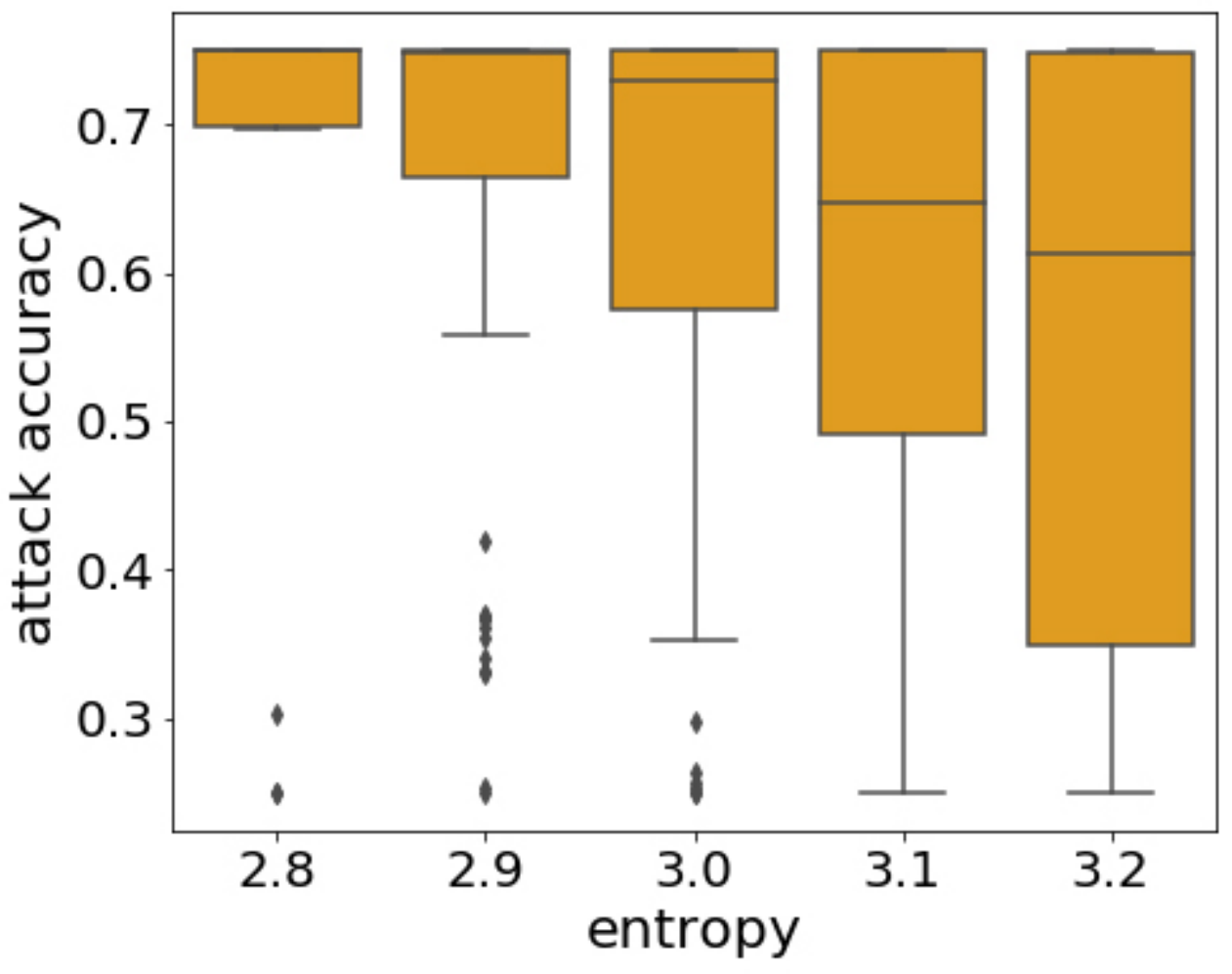}
    }
    \hfill
    \subfloat[Adult dataset.]{
        \includegraphics[width=.29\textwidth]{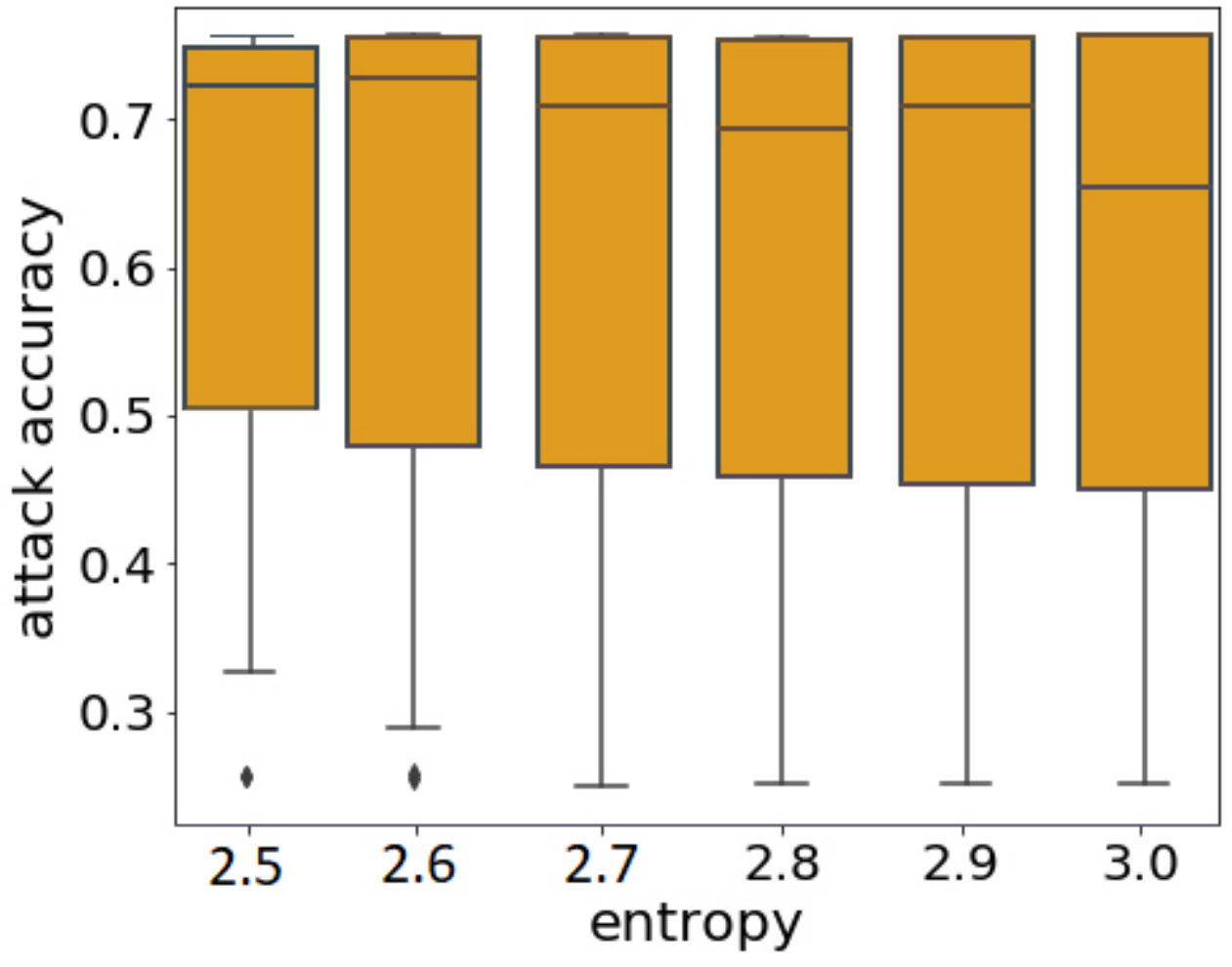}
    }
    \caption{\zhigang{Effects of the entropy of the training dataset on MIA accuracy.} 
    }
    \label{fig:p_en_attack}
\end{figure*}

\textbf{The Entropy of The Training Dataset:}
~Figure \ref{fig:p_en_attack} shows the attack accuracy as a function of entropy of the training dataset. 
%
In general, as the entropy of the training dataset increases,the MIA accuracy decreases. This is particularly true for the Purchase and Texas datasets. The decrease is slower for the Adult dataset. 
This negative impact of the entropy on the attack accuracy is due to higher amount of randomness increasing the difficulty to infer membership information from the dataset. 



\subsection{Effect of Model Properties on MIA}\label{sec:model_prop}
\textbf{Model and Training Parameters.} 
We assess whether model and training parameters can impact the MIA accuracy. For the model parameters, we vary the depth of the ANN model from 1 to 5 hidden layers and the number of nodes of each layer from 5 to 1000. 

For the training parameters, we vary the learning rate $\alpha$ (from $10^{-5}$ to $1$) and the L2-ratio regularizer $\lambda$ (between 0 and 3). Then we ensure that all configurations tested against MIA provide similar prediction performance for the ML model, e.g., by pruning off those with lower ($<0.9$) training and test accuracy. 

\begin{figure*}[ht!]
\centering
\captionsetup{justification=centering}
    \subfloat[Purchase dataset.]{
        \includegraphics[width=\textwidth]{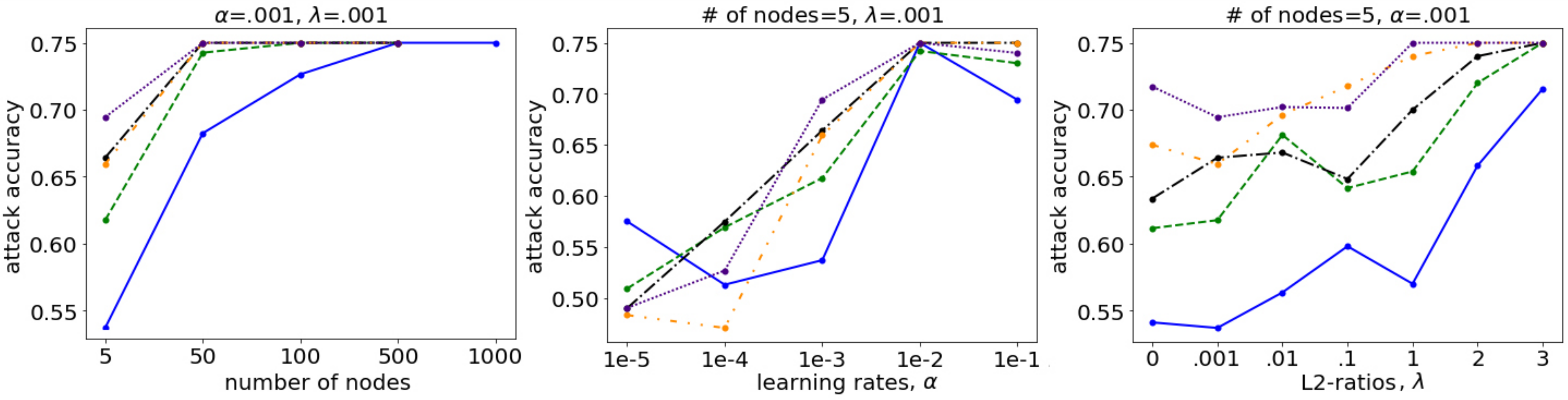}
    }
    \hfill
\subfloat[Texas dataset.]{
        \includegraphics[width=\textwidth]{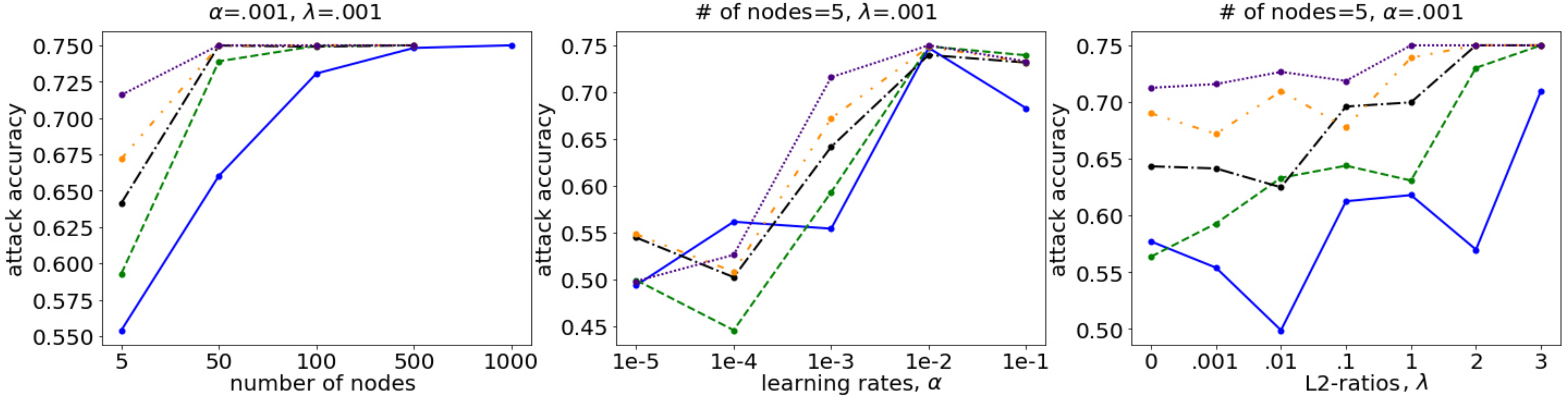}
    }
    \hfill
\subfloat[Adult dataset.]{
        \includegraphics[width=\textwidth]{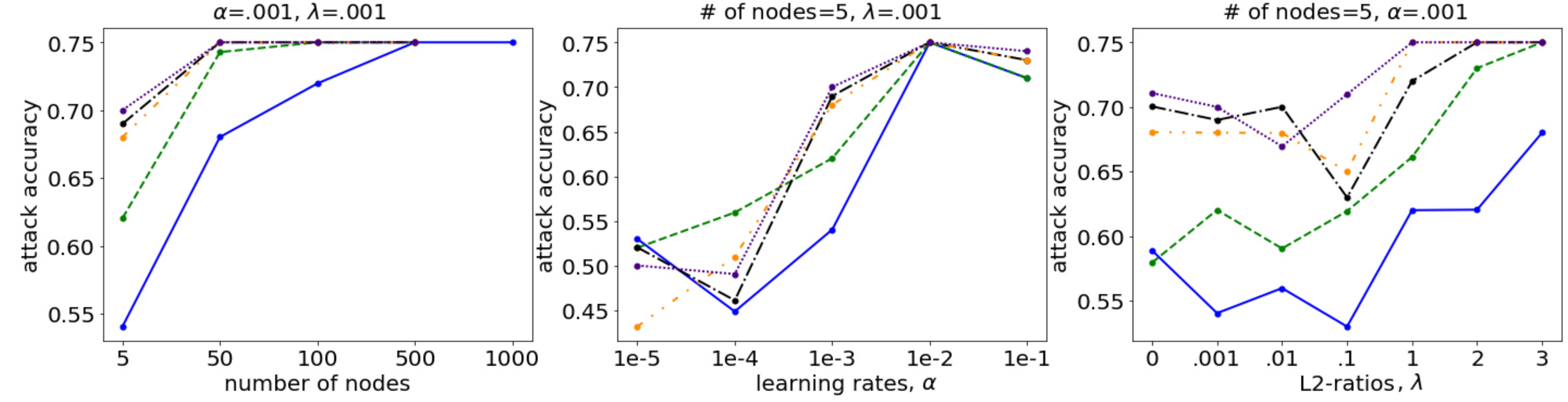}
    }
    \hfill
\captionsetup[subfigure]{labelformat=empty}
\centering
\subfloat{
        \includegraphics[width=.5\textwidth]{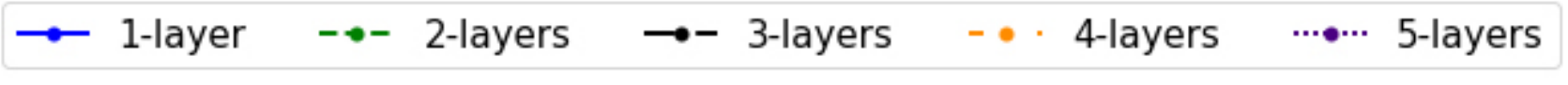}
    }
    \hfill
\caption{MIA accuracy for the i) different number of nodes, ii) different learning rates and iii) different L2-ratios \zhigang{over different datasets}. 
}
\label{fig:hyper_attack}
\vspace{-4mm}
\end{figure*}

Figure~\ref{fig:hyper_attack} reports the MIA accuracy obtained for different numbers of layer nodes and different training parameters $\alpha$ and $\lambda$ using 1 to 5 hidden layers.  The results show that \textit{deeper} ANN architectures (e.g., 5 hidden layers) are consistently more vulnerable to MIA compared to shallower configurations (e.g., 1 layer only). We also observe that the higher the number of nodes in the model the higher the MIA accuracy.


Furthermore, Figure~\ref{fig:hyper_attack} shows that the choice of the training parameters has a significant impact on MIA accuracy. In particular, we observe that large learning rates (e.g., 0.1) result in more accurate MIA attacks. Concerning the effect of the L2-ratio regularizer, from $\lambda=0$ onwards we observe a decrease of MIA accuracy. However, after a certain point (e.g., $\lambda=0.1$ for a 1-layer ANN on Adult dataset), increasing the amount of regularization makes MIAs more effective than in the case of no regularization ($\lambda=0$).




\begin{table}[ht!]
    \centering
    \captionsetup{justification=centering}
    \subfloat[Purchase dataset.]{
        \resizebox{.4\textwidth}{!}{
        \begin{tabular*}{.5\textwidth}{ccccccc}
        \multirow{2}{*}{Target models}& \multicolumn{6}{c}{Shadow models} \\ \cmidrule(l){2-7}
        & ANN & $k$NN & LR & RF & SVM & All \\ \hline
        ANN & \textbf{60.09} & 52.38 & 58.11 & 55.41 & 52.38 & 69.25 \\
        $k$NN & 55.04 & 55.92 & \textbf{55.57} & 53.21 & 51.68 & 70.45 \\
        LR & 51.14 & 53.55 & 54.08 & 56.51 & \textbf{57.23} & 70.49 \\
        RF & 56.26 & 55.16 & 54.07 & \textbf{57.69} & 50.93 & 71.84 \\
        SVM & 51.09 & \textbf{58.56} & 55.69 & 51.61 & 52.99 & 70.11 \\
        \end{tabular*}}
    }
\hfill
    \subfloat[Texas dataset.]{
        \resizebox{.4\textwidth}{!}{
        \begin{tabular*}{.5\textwidth}{ccccccc}
        \multirow{2}{*}{Target models} & \multicolumn{6}{c}{Shadow models} \\ \cmidrule(l){2-7} 
        & ANN & $k$NN & LR & RF & SVM & All \\ \hline
        ANN & 57.62 & 57.07 & 54.34 & \textbf{61.23} & 56.54 & 69.97 \\
        $k$NN & 55.30 & 54.57 & 54.36 & \textbf{57.52} & 51.37 & 68.45 \\
        LR & 52.08 & 54.69 & \textbf{55.82} & 54.09 & 55.29 & 67.54 \\
        RF & 53.88 & 55.78 & 58.90 & 57.11 & \textbf{59.61} & 72.64 \\
        SVM & 56.43 & 50.62 & 53.94 & 52.35 & \textbf{57.14} & 62.88 \\
        \end{tabular*}}
    }
\hfill
    \subfloat[Adult dataset.]{
        \resizebox{.4\textwidth}{!}{
        \begin{tabular*}{.5\textwidth}{ccccccc}
        \multirow{2}{*}{Target models} & \multicolumn{6}{c}{Shadow models} \\ \cmidrule(l){2-7} 
        & ANN & $k$NN & LR & RF & SVM & All \\ \hline
        ANN & \textbf{59.76} & 56.76 & 53.30 & 52.27 & 55.56 & 70.28 \\
        $k$NN & 50.95 & \textbf{56.42} & 51.97 & 50.57 & 54.62 & 70.98 \\
        LR & 56.18 & 53.68 & 55.04 & \textbf{56.97} & 47.47 & 70.18 \\
        RF & 55.43 & \textbf{60.09} & 46.88 & 59.27 & 52.69 & 72.76 \\
        SVM & \textbf{61.45} & 53.83 & 51.82 & 55.02 & 52.74 & 73.06 \\
        \end{tabular*}}
    }
\caption{MIA accuracy for the different target and shadow model combinations. The right-most column ("All") shows the MIA accuracy for each of the target models against all the five examined ML algorithms used as the shadow models.}
\label{tab:model_model}
\end{table}

\textbf{Target-Shadow Model Combination.} 
~Table \ref{tab:model_model} reports the MIA accuracy for different target and shadow model types. Specifically, we test different ML classifiers, including $k$-Nearest Neighbors ($k$NN), Logistic Regression (LR), Random Forest (RF) and Support Vector Machine (SVM). We observe that different combinations of target and shadow model architectures generally result in different MIA accuracy, and that the most effective shadow-model choice also varies across different datasets. Interestingly, using shadow models with an architecture type that matches the target does not guarantee maximum MIA accuracy. Overall, these results support the notion of MIA \textit{transferability} introduced in~\cite{truex2019effects}, where attackers do not always need to know the exact target model configuration to launch an effective MIA as attack models can be transferable from one target model type to another.

In addition, we evaluate the MIA accuracy in a  \textit{one-versus-all} scenario, where each target model architecture is consecutively tested against the five different shadow model architectures. From Table \ref{tab:model_model}, we observe in this new scenario a significant increase (in the range of +10\%) of MIA accuracy compared to previous \textit{one-to-one} cases. We can conclude that the MIA accuracy is highly classifier-dependent, and that the combined use of different shadow model architectures results in much more severe attacks. 


\textbf{Mutual Information between Records and Model Parameters.}
~\gio{
The mutual information $I(X;\theta)$ between the ML model parameters and the training records, defined in Section~\ref{sec:data}, is a measure of the information \textit{learnt} by the model over the dataset, i.e., captured from the features of the training records. Higher values of $I(X;\theta)$ may be as such indicative of increased model's vulnerability in the face of MIAs. To experiment with different levels of mutual information, we \shakila{use the default ANN model} 
and re-train it from scratch 1000 times \zhigang{(we perform 200 experiments for $10\%$ to $50\%$ class balances)} 
 with random parameter initialization. At each iteration, we collect all model parameters and derive $I(X;\theta)$. Lastly, we group the different model instances based on the $I(X;\theta)$ range.} 

\begin{figure*}[ht!]
    \centering
    \captionsetup{justification=centering}
    \subfloat[Purchase dataset.]{
        \includegraphics[width=.29\textwidth]{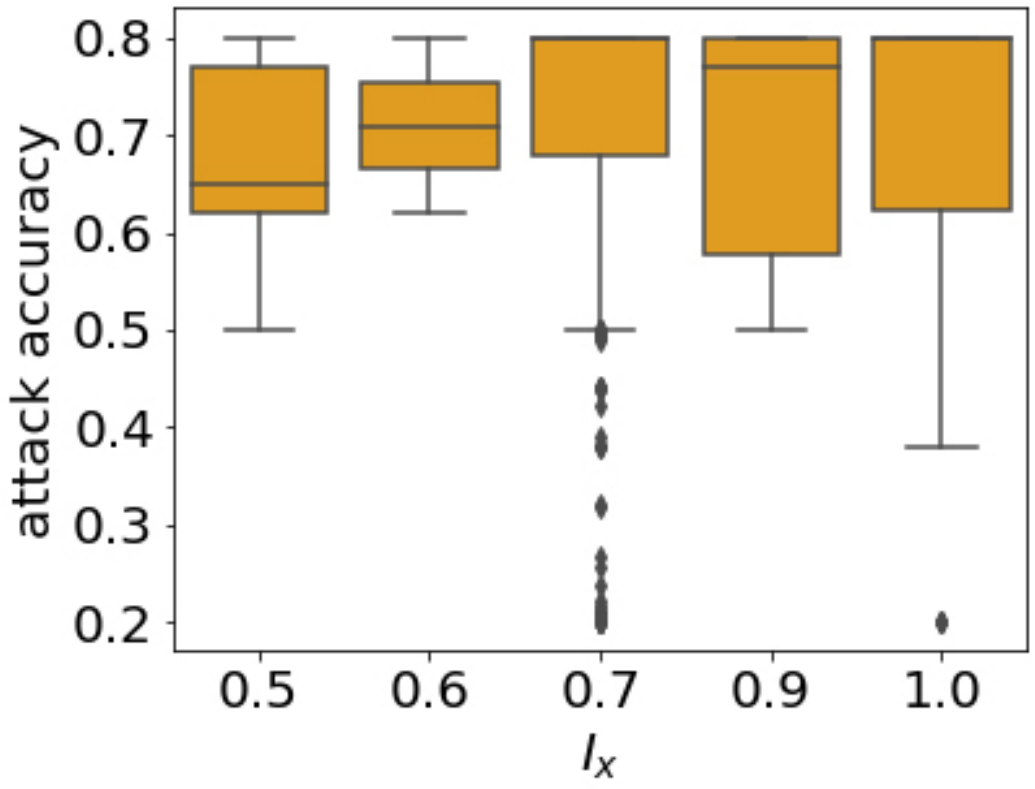}
    }
    \hfill
    \subfloat[Texas dataset.]{
        \includegraphics[width=.29\textwidth]{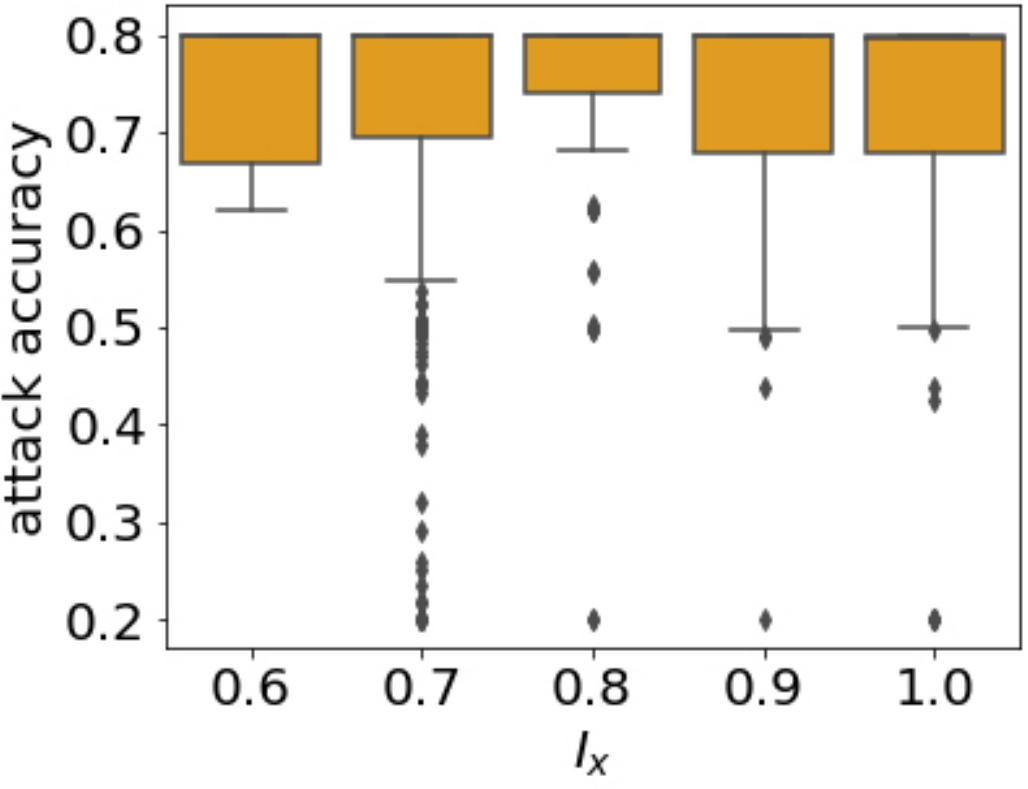}
    }
    \hfill
    \subfloat[Adult dataset.]{
        \includegraphics[width=.29\textwidth]{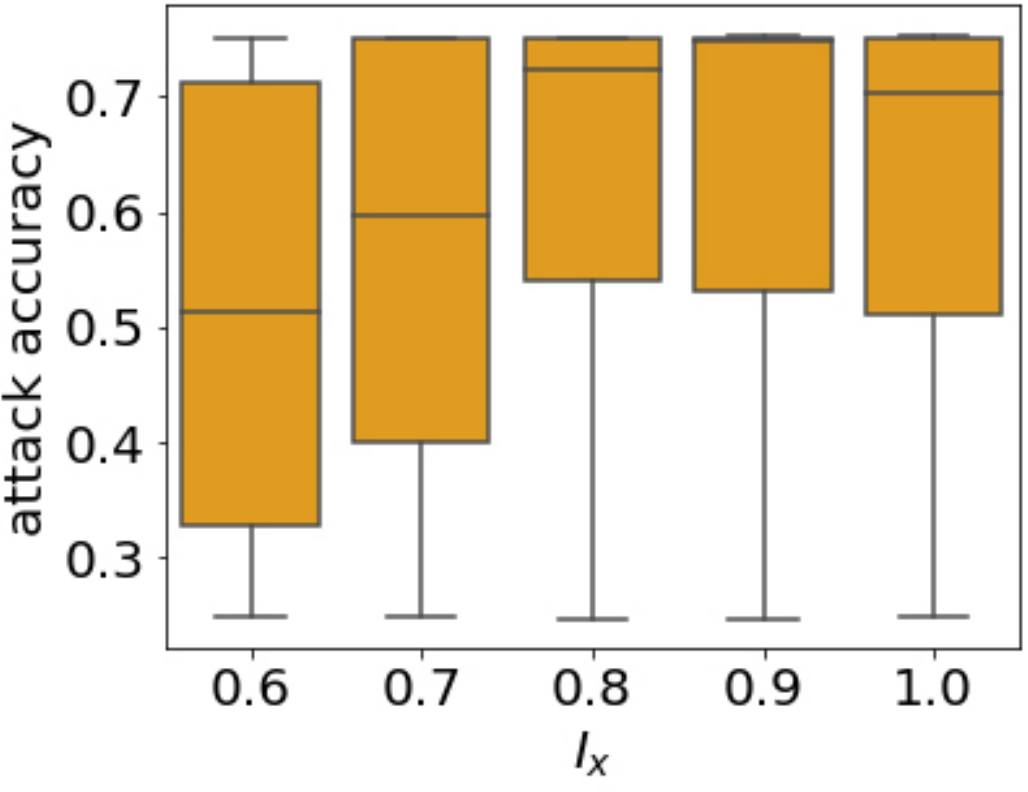}
    }
    \caption{Relationship between the Mutual Information, 
    $I_X$ and MIA accuracy.
    }
    \label{fig:p_mi_attack}
\end{figure*}

The results shown in Figure \ref{fig:p_mi_attack} confirm  that higher values of $I(X;\theta)$ may indicate increasing attack's exposure to MIA. In particular, despite some differences in the results between datasets, we observe that higher mutual information is generally associated with more accurate MIA. 

We conjecture that the observed difference between the Adult dataset and the other datasets in terms of MIA accuracy ranges, comes from the lower data size (10K instead of 100K). Even if the values of average mutual information $I(X;\theta)$ are the same across datasets, the smaller training set in the case of the Adult dataset results in less total information ``learnt'' by the model during training, which in turn reduces the MIA accuracy. 

\begin{figure*}[ht!]
    \centering
    \captionsetup{justification=centering}
    \subfloat[Purchase dataset.]{
        \includegraphics[width=.29\textwidth]{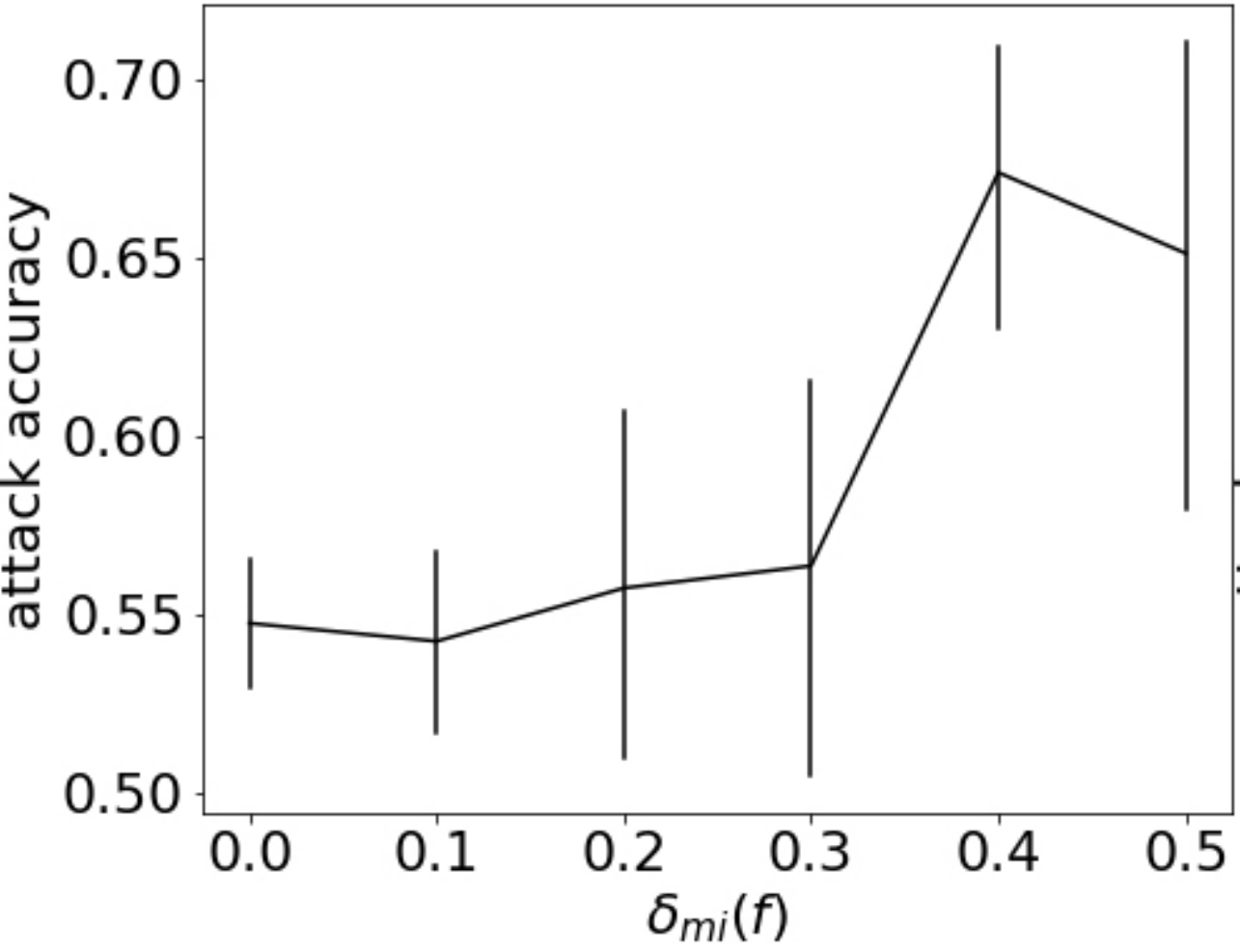}
    }
    \hfill
    \subfloat[Texas dataset.]{
        \includegraphics[width=.29\textwidth]{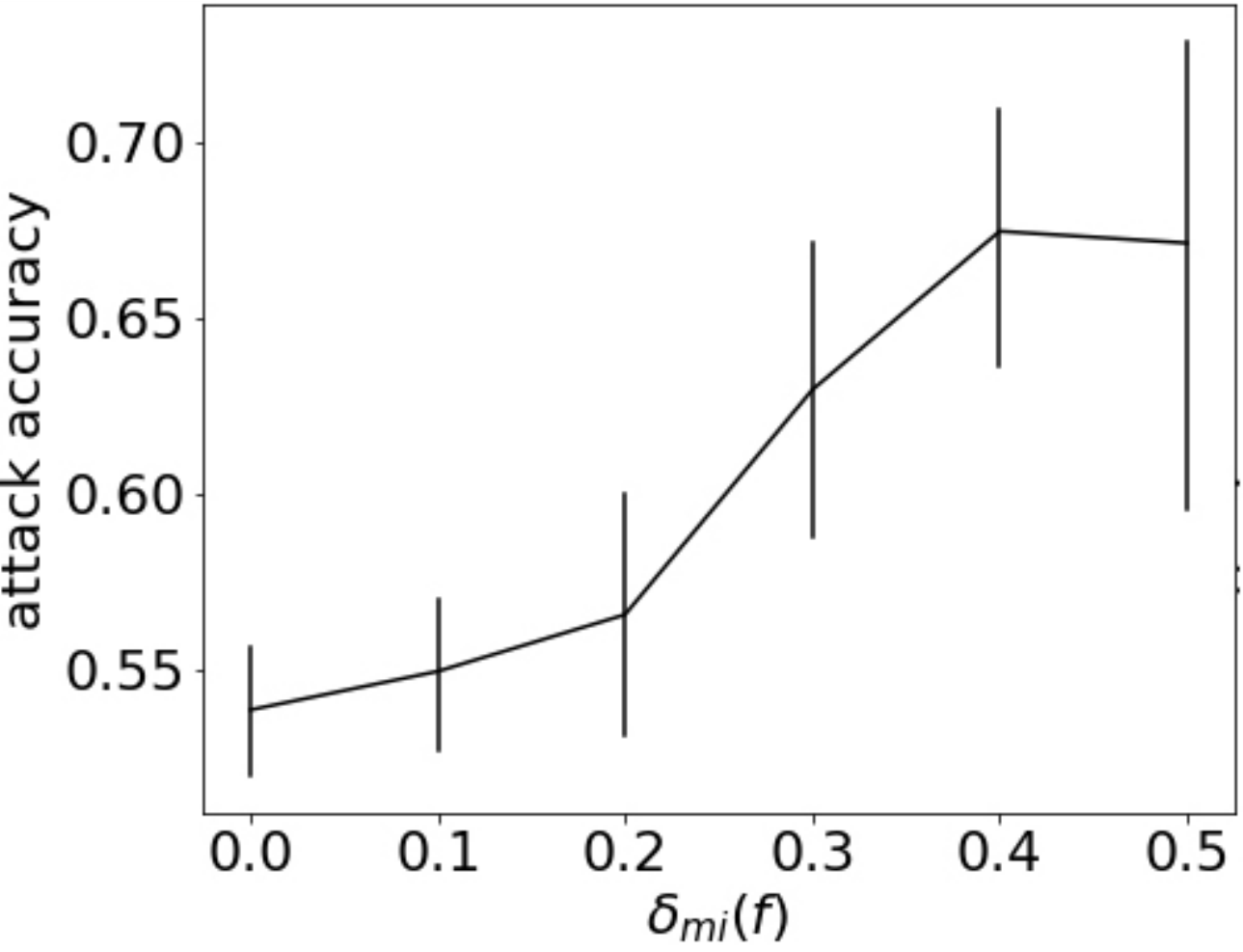}
    }
    \hfill
    \subfloat[Adult dataset.]{
        \includegraphics[width=.29\textwidth]{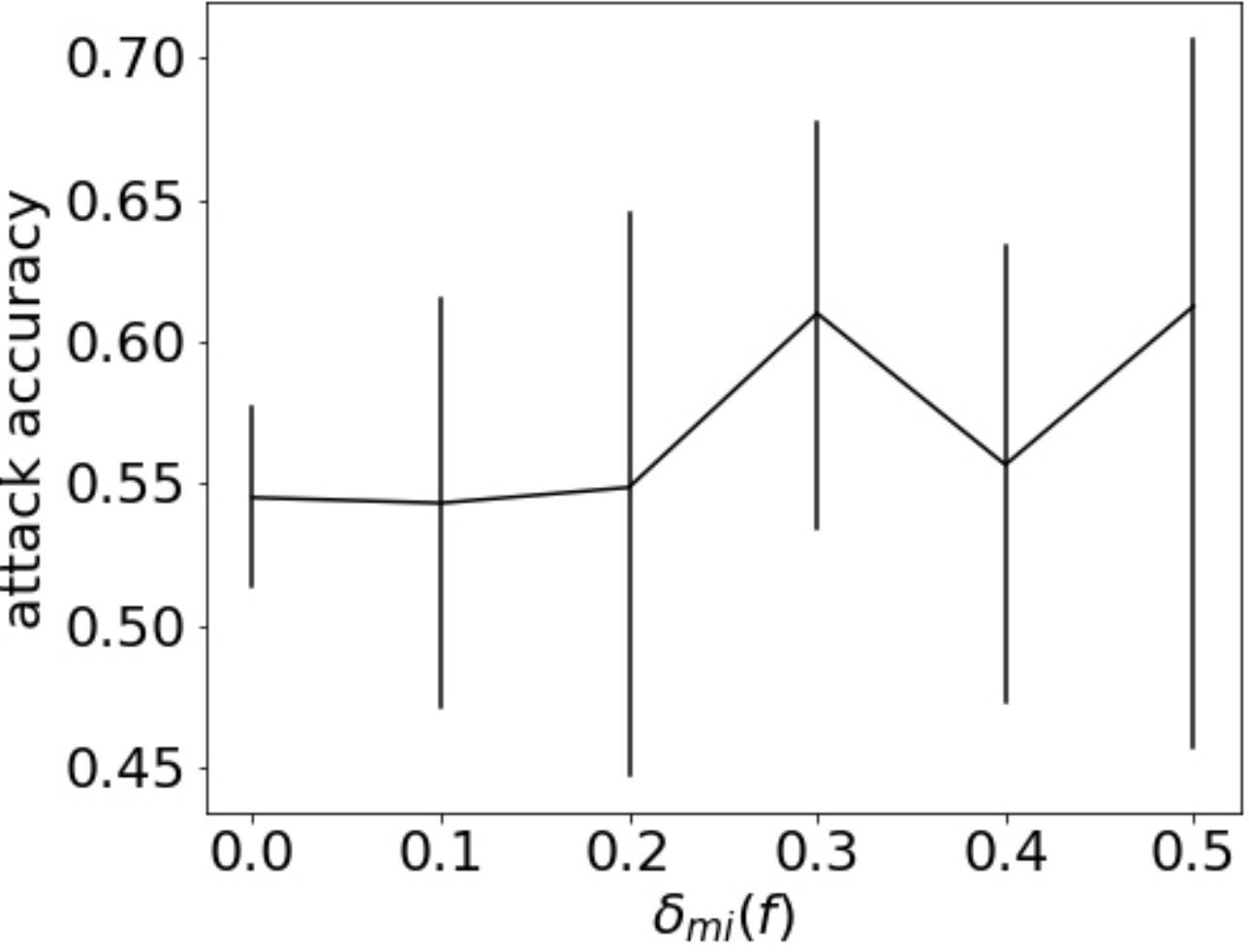}
    }
    \caption{Results showing the level of a model's vulnerability towards MIA as the deviation from the perfect MIA-indinguishability against MIA accuracy.} 
    \label{fig:p_member_attack}
\end{figure*}

\textbf{MIA-indistinguishability.}  
~\gio{The \textit{perfect MIA-indistinguishability} property introduced in ~\cite{yaghini2019disparate} (Eq.\ref{eq:mem_non}) indicates perfect resistance of a ML model to MIAs. Here we analyze the relation between the deviation of the ML model from perfect MIA-indistinguishability $\delta_{mi}(f)$ and its vulnerability to MIA. As described in Section~\ref{sec:data}, $\delta_{mi}(f)$ is measured as the difference between the training-set member and non-member prediction probabilities. Intuitively, we expect lower values of $\delta_{mi}(f)$ to result in limited MIA accuracy, and larger deviations to enable more powerful attacks. To experiment with different ranges of $\delta_{mi}(f)$, we enforce different sampling biases on the training data, especially by varying the train/test split and the class balance.

Figure \ref{fig:p_member_attack} shows that while small deviations (0 to 0.1) from perfect MIA-indistinguishability generally guarantee low MIA accuracy (50 to 55\% MIA accuracy), increasing deviations not always result in more successful attacks. This is particularly evident in the case of Adult dataset, where the MIA accuracy falls in very similar ranges for considerably different values of $\delta_{mi}(f)$ such as 0.1 and 0.4.} 

\begin{figure*}[ht!]
\centering
\captionsetup{justification=centering}
\subfloat[Group fairness.]{
        \includegraphics[width=.29\textwidth, height=.25\textwidth]{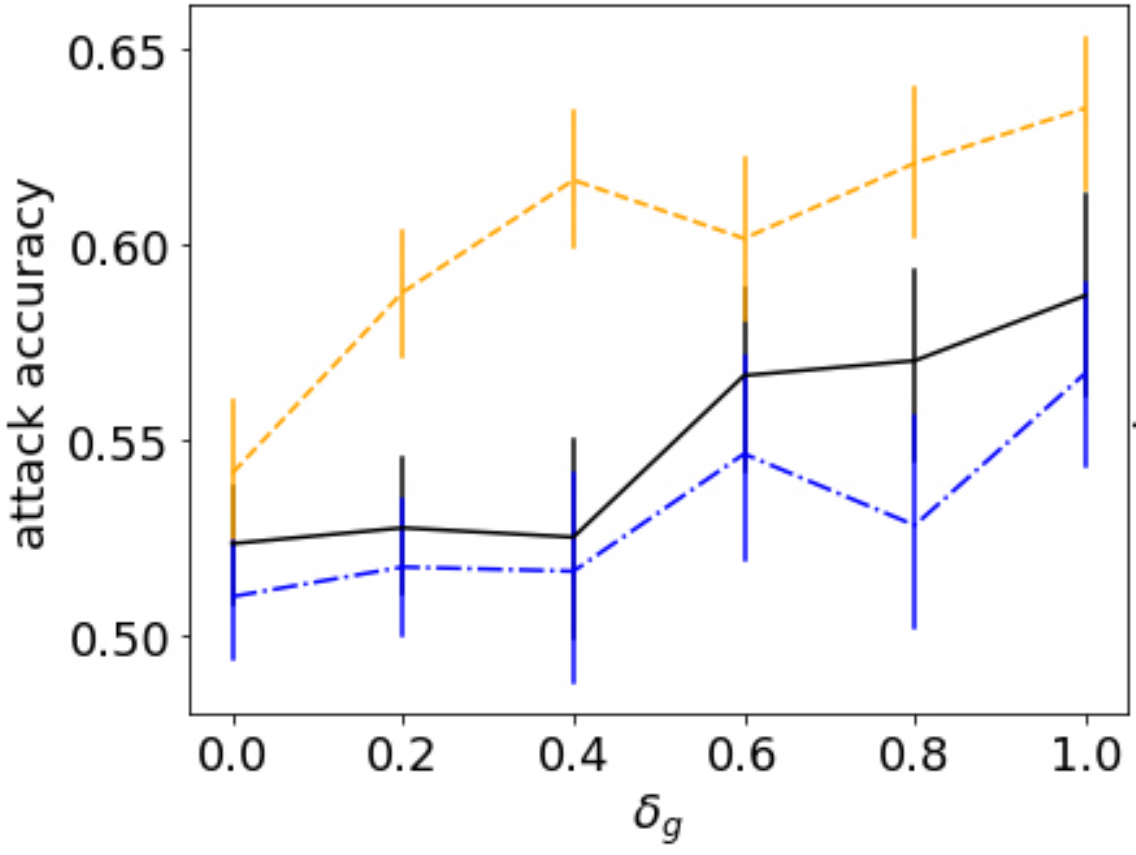}
    }
    \hfill
\subfloat[Predictive fairness.]{
        \includegraphics[width=.29\textwidth, height=.25\textwidth]{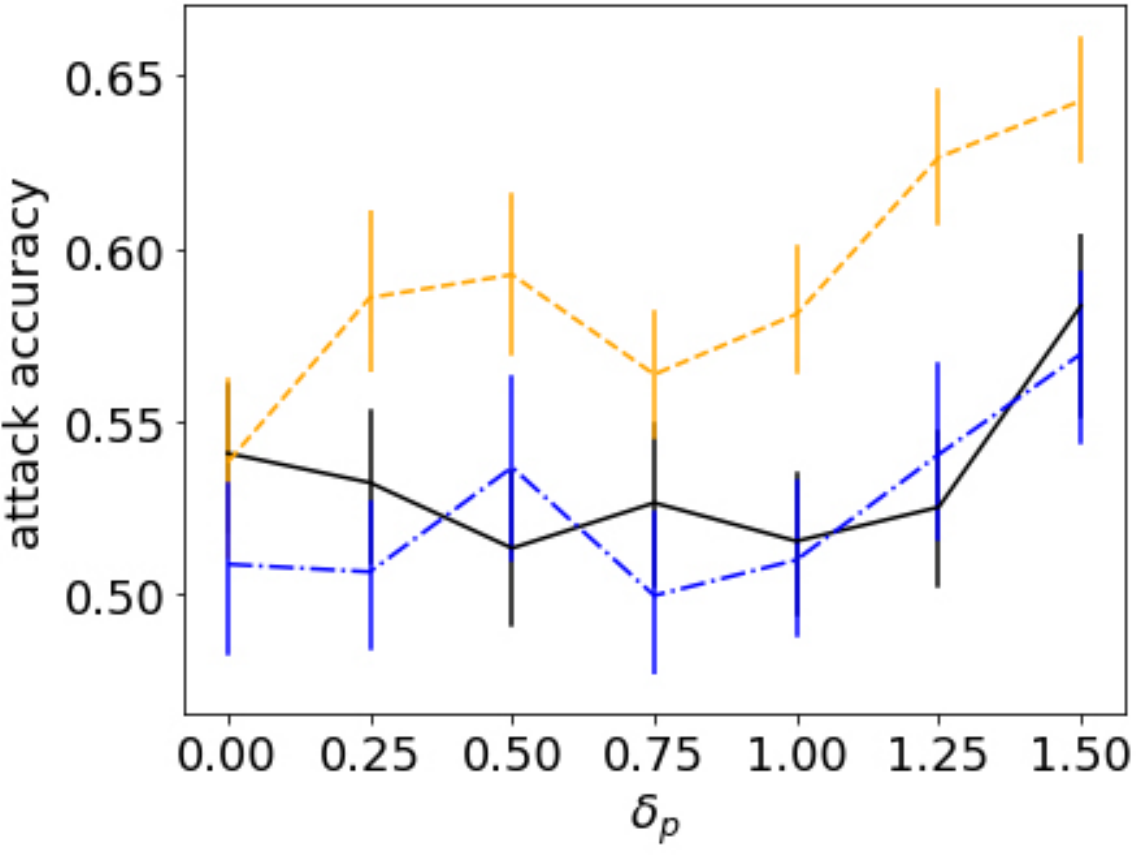}
    }
    \hfill
\subfloat[Individual fairness.]{
        \includegraphics[width=.29\textwidth, height=.25\textwidth]{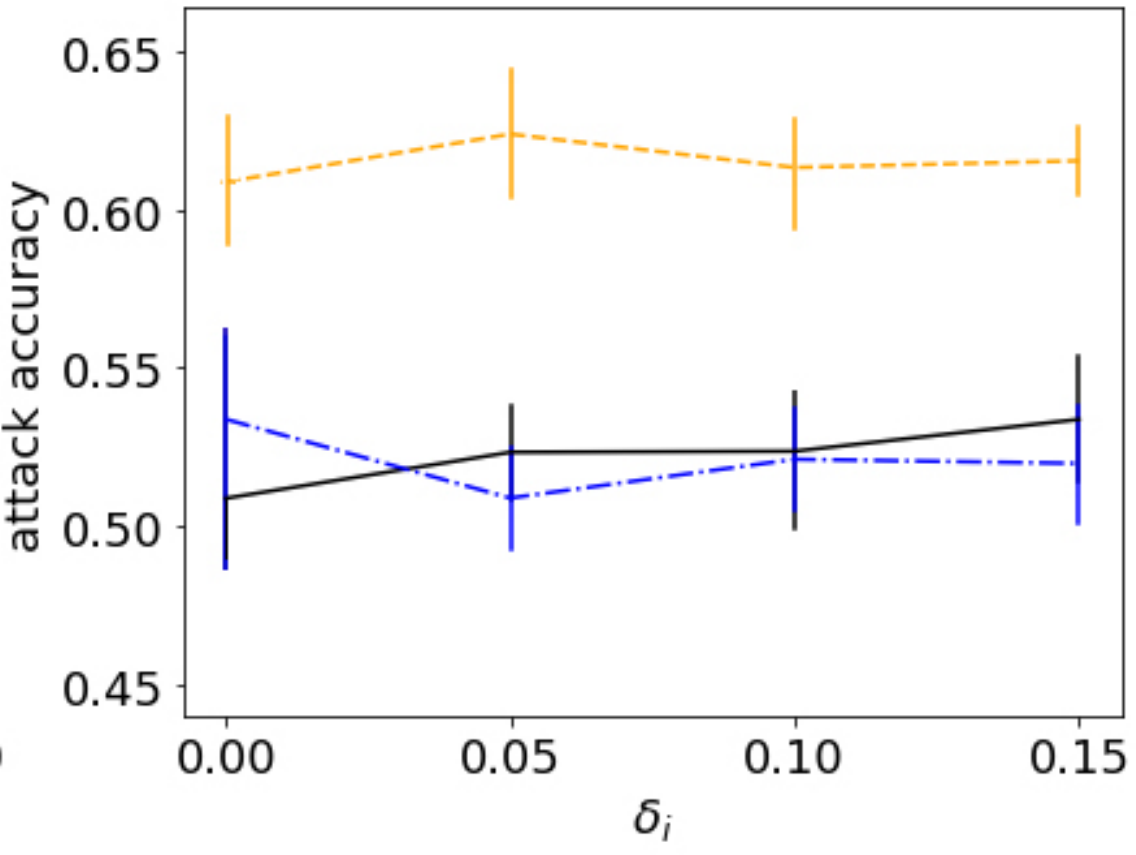}
    }
    \hfill
\subfloat{
        \includegraphics[width=.5\textwidth]{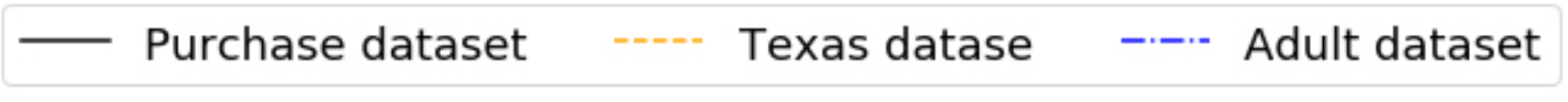}
    }
    \hfill
\caption{Impact of model's fairness on MIA accuracy. Larger values of $\delta_g$, $\delta_p$, $\delta_i$ indicate reduced prediction fairness. 
Data size is 100,000, number of features is 5, class and feature balances are below 50\%.}
\label{fig:p_fairness_attack}
\end{figure*}

\textbf{Model's Fairness.} 
~\gio{The fairness metrics defined in Section~\ref{sec:data} indicate the difference of the model's prediction output between individual dataset records or groups of records, with smaller values of these metrics representing fairer model behaviors. Here we analyze the impact of group, predictive, and individual fairness on the MIA-vulnerability by assessing the extent to which \textit{unfair} model behaviors result in higher MIA accuracy. To experiment with different fairness levels, we build on the empirical observation that  more balanced datasets determine fairer prediction behaviors. Hence, we vary the class and feature balance of the data and for each configuration me measure both the model fariness and the MIA accuracy, thus obtaining the results in Figure~\ref{fig:p_fairness_attack}. These results show reduced MIA accuracy when fairness is high, \textit{i.e.}, $\delta_g,\delta_p,\delta_i$ are close to 0. However, not all fairness properties have the same effect on MIA; for example group fairness shows a stronger and more consistent impact on MIA accuracy across different datasets compared to individual fairness.}

\begin{table*}[t]
    \centering
    \captionsetup{justification=centering}
    \subfloat[1-layer, no regularizer ($\lambda$=0).]{
        \resizebox{.35\textwidth}{!}{
            \begin{tabular}{lcccc}
            Datasets & Train Acc. & Test Acc. & MIA Acc. \\\hline
            Adult & 80.956 & 81.232 & 58.893 \\
            Purchase & 71.512 & 71.405 & 54.121 \\
            Texas & 78.465 & 78.226 & 57.716 \\
            \end{tabular}
        }}
    \hspace{3.5mm}
    \subfloat[5-layers, no regularizer).]{
        \resizebox{.35\textwidth}{!}{
            \begin{tabular}{lccc}
            Datasets & Train Acc. & Test Acc. & MIA Acc. \\\hline
            Adult & 89.806 & 89.782 & 71.080 \\
            Purchase &88.802  & 88.780 & 71.736 \\
            Texas & 90.993  & 90.862 & 71.257 \\
        \end{tabular}
    }}
    \\
    \subfloat[1-layer, L2 regularizer ($\lambda$=0.01).]{
        \resizebox{.35\textwidth}{!}{
            \begin{tabular}{lccc}
            Datasets & Train Acc. & Test Acc. & MIA Acc. \\\hline
            Adult & 74.972 & 75.076 & 55.977 \\
            Purchase & 64.829 & 65.532 & 56.36 \\
            Texas & 75.974 & 76.074 & 49.891 \\
            \end{tabular}
    }}
    \hspace{3.5mm}
    \subfloat[5-layers, L2 regularizer ($\lambda$=0.01).]{
        \resizebox{.35\textwidth}{!}{
            \begin{tabular}{lccc}
                Datasets & Train Acc. & Test Acc. & MIA Acc. \\\hline
                Adult & 88.364 & 88.355 & 66.955 \\
                Purchase &88.33  &88.358 & 70.211 \\
                Texas & 90.498 & 90.473 & 72.664 \\ 
            \end{tabular}
    }}
\caption{Train, test and MIA accuracy (\%) obtained on the target models having 1 and 5 hidden layers with $\lambda = 0.0$ and $\lambda = 0.01$.
}  
\label{tab:overfit}
\end{table*}

\subsection{Discussion}


\gio{
\textbf{Impact of different data and model properties on MIA.}
~Our experimental analyses found that a variety of properties have an effect on MIA, which concern the data, the target (and shadow) model architecture, as well as the model prediction behavior. About the data, we observed that the shadow data size (volume of the attacker's data), the class \zhigang{and feature} balance in the target-model training set \zhigang{, and the entropy of the training dataset} are the most influential factors, \gio{producing variations of MIA accuracy between of 0.5 and 0.75.} 
Concerning the ML model, we found that ML model designs or settings that are equivalent in terms of prediction performance (i.e., similar train/test accuracy), can have considerably different exposures to MIAs. This depends especially on the model architecture and hyperparameters (e.g., deeper and larger neural networks appear much more MIA-vulnerable) and on the model's prediction fairness across subgroups of dataset records, which can \shakila{increase or decrease} the MIA accuracy respectively by up to 0.2 
for all datasets. 


For other properties, such as the size and the number of features in the model training set, the relation with MIA is weaker as no clear trend could be identified from the experiments. 
Furthermore, our results have shown that not only different properties have a different effect on MIA, but the impact of a specific property can also be strongly dataset-dependent. This is evident for properties such as the dataset entropy and the mutual information $I_X$, whose impact on MIA is substantially different for the Adult dataset compared to the other two. Cases like this require further exploration, in particular for unveiling the role of individual feature on the MIA, e.g., looking at the distance between individual records among their feature values~\cite{zhao2019inferring}.

\textbf{Overfitting does not explain it all.} 
~Recent research ~\cite{nasr2018machine, 10shokri2017membership, truex2019effects} pointed to overfitting as the main contributor to MIA success. Interestingly, all the different MIA accuracy values shown in our analyses are obtained using non-overfitted target models. These results are in accordance with \cite{long2018understanding}, and confirm that overfitting is by no means necessary for exposing membership information from training data. To further illustrate this, Table \ref{tab:overfit} reports the MIA accuracy along with the train and test accuracy of the target model measured for different hyperparameter and training settings (specifically, the number of layers and the L2 regularizer). For each dataset, we observe significant differences in MIA accuracy across the four sub-tables, despite none of these configurations presents any overfitting -- all settings produce very similar train and test accuracy. 

\textbf{Hard to identify the main property responsible for MIA's success.}
~Overall, the results in Section~~\ref{sec:data_prop} and Section~\ref{sec:model_prop} show that a range of different properties contribute to MIA success. Finding which one is the main property that is responsible for the MIA-vulnerability of a ML model is a particularly challenging task, given also that data/model properties are often interdependent. For example, properties such as the class and feature balance are correlated with each other, and the same applies to the different measures of prediction fairness. To better visualize these relations, Table~\ref{tab:interrelations} presents both the linear (Pearson's) and rank (Spearman's) correlation coefficients between pairs of data and model properties from Sections~~\ref{sec:data_prop} and~\ref{sec:model_prop}. The prevalence of 0.3 to 0.7 (-0.3 to -0.7) values indicates that most of the pairs have at least a moderate positive (or negative) relationship. \shakila{Furthermore, Table~\ref{tab:interrelations} shows that model properties can also be correlated to data properties. For example, the dataset class balance is related to the predictive fairness of a ML model, and the same happens between group or individual fairness and the feature balance.} 


\textbf{No property can alone measure the exposure of ML models to MIA.}
~A further consequence of the fact that multiple properties affect the MIA performance is that none of them can alone precisely measure the model resilience against membership inference. Overfitting, as shown above, does not provide an ultimate indication of MIA-vulnerability as  non-overfitted models can be heavily exposed to the attack. Similar considerations apply to MIA-indistinguishability, for which we found that larger deviations from perfect MIA-indistinguishability ($\delta_{mi}(f)$ in Figure~\ref{fig:p_member_attack}) not always reflect on more MIA-vulnerable ML models as observed across different datasets. Nonetheless, due to their demonstrated relation with MIA accuracy, several properties can still be exploited when developing defense methods for ML models against MIAs. We explore these opportunities in the next section. 
}

\begin{table}[ht!]
\centering
    \captionsetup{justification=centering}
    \subfloat[Purchase dataset.]{
    \resizebox{.45\textwidth}{!}{%
    \begin{tabular}{@{}llcc@{}}

    Prop. 1 & Prop. 2 & Pearson's r & Spearman's r \\ \midrule
    Class Balance & Feature Balance & -0.329 & -0.709 \\
    Class Balance & Entropy & 0.473 & 0.291 \\
    Feature Balance & Entropy & 0.541 & 0.367 \\
    Group Fair. & Predictive Fair. & 0.24  & 0.2 \\
    Group Fair. & Individual Fair. & 0.537  & 0.461 \\
    Predictive Fair. & Individual Fair. & 0.39 & 0.38 \\ 
    Group Fair.&	Class Balance&	-0.185&	-0.227\\
    Predictive Fair.&	Class Balance&	-0.777&	-0.793\\
    Individual Fair.&	Class Balance&	-0.001&		0.001\\
    Group Fair.&	Feature Balance&	-0.588&	-0.465\\
    Predictive Fair.&	Feature Balance&	-0.398&		-0.391\\
    Individual Fair.&	Feature Balance&	-0.972&	-0.979\\
\end{tabular}}
}
\hfill
\subfloat[Texas dataset.]{
    \resizebox{.45\textwidth}{!}{%
    \begin{tabular}{@{}llcc@{}}

    Prop. 1 & Prop. 2 & Pearson's r & Spearman's r \\ \midrule
    Class Balance & Feature Balance & -0.831	& -0.955 \\
    Class Balance & Entropy &  0.972	& 0.994   \\
    Feature Balance & Entropy&  -0.801	&-0.944   \\
    Group Fair. & Predictive Fair.& 0.425	&0.421  \\
    Group Fair. & Individual Fair. &  0.552 &	0.322  \\
    Predictive Fair. & Individual Fair. &  0.194 &	0.088   \\ 
    Group Fair.&	Class Balance&  -0.17 &	-0.179  \\
    Predictive Fair.&	Class Balance&  -0.703 &	-0.719  \\
    Individual Fair.&	Class Balance&  0.002 &	0.002  \\
    Group Fair.&	Feature Balance&  -0.818	& -0.787  \\
    Predictive Fair.&	Feature Balance&  -0.541	&-0.533  \\
    Individual Fair.&	Feature Balance& -0.441	&-0.292 \\
\end{tabular}}
}
\hfill
\subfloat[Adult dataset.]{
\resizebox{.45\textwidth}{!}{%
    \begin{tabular}{@{}llcc@{}}

    Prop. 1 & Prop. 2 & Pearson's r & Spearman's r \\ \midrule
    Class Balance & Feature Balance &  -0.962 & 	-0.965 \\
    Class Balance & Entropy &  0.833 & 	0.817 \\
    Feature Balance & Entropy&  -0.825& 	-0.815  \\
    Group Fair. & Predictive Fair.&  0.104& 	0.056  \\
    Group Fair. & Individual Fair. &  0.378& 	0.165  \\
    Predictive Fair. & Individual Fair. &  0.164& 	0.11  \\ 
    Group Fair.&	Class Balance&  -0.178& 	-0.215 \\
    Predictive Fair.&	Class Balance& -0.682& 	-0.699  \\
    Individual Fair.&	Class Balance&  -0.007& 	-0.014  \\
    Group Fair.&	Feature Balance&  -0.632& 	-0.523  \\
    Predictive Fair.&	Feature Balance&  -0.408& 	-0.401 \\
    Individual Fair.&	Feature Balance&  -0.418& 	-0.274 \\
\end{tabular}}
}
\caption{\gio{Pearson's and Spearman's correlation coefficient between different properties. In all the cases the coefficients are statistically significant (p-value < 0.001).}}
\label{tab:interrelations}
\end{table}

\section{Towards  MIA-resilient ML models}\label{sec:defense}





\zhigang{Based on the experimental results in Section~\ref{sec:result}, we propose to use model properties} \gio{that} \zhigang{positively (or moderately positively) impact MIA accuracy as regularizers for the ML model training. It is intuitive that minimising such a regularizer can reduce MIA accuracy while improving the underlying model's performance. Algorithm~\ref{alg:MIA_defense} depicts our main idea where \shakila{for $k$ number of epochs,} we calculate a given fairness or the mutual information between the records and model parameters \shakila{$l$ number of times}in Line 3-7, then use it as the regularizer for model training in Line 9, where $l(\cdot)$ is the loss function, $\sum_{j}|\theta \times e^{r}|$ is the regularizing function with weight $\lambda$. } 

\begin{algorithm}[t!] 
    \DontPrintSemicolon
    \SetKwFor{For}{for}{do}{end~for}
    \KwData{
    \zhigang{$D(X, y)$: target training dataset; $k, l, m$: constants.}
    }
    \zhigang{$g_{0}$, $g_{1}$ $\gets$ $g_{0} \cup g_{1} = D$, $g_{0} \cap g_{1} = \phi$}\;
    \For{$k$ number of epochs}{
        \For{$l$ times}{
            \zhigang{$D^{\prime} \gets$ sampling $m$ records $(x_{i}, y_{i})$ from $D$}\;
            \zhigang{$f_{t}, \theta \gets$ target model trained on $D^{\prime}$}\;
            \zhigang{$\hat{y} \gets$ prediction results on $g_{0}, g_{1}$}\;
            \zhigang{$r \gets$ choose one from $\{\delta_{g}(f_{t},\hat{y}), \delta_{p}(f_{t},\hat{y}), \delta_{i}(f_{t},\hat{y}), I(X;\theta)\}$}\;
        }
        \zhigang{$r \gets \max_{l}r$}\;
        \zhigang{$f_{t} \gets$ descend gradients over $\theta$ by}
        $$\bigtriangledown_\theta \frac{1}{m}\sum_{i=1}^{m}l(f_t(x_i),y_i) + \lambda \sum_{j}|\theta \times e^r|$$\;
        \vspace{-7mm}
    }
    \caption{Training a model using \zhigang{our proposed} regularizer.}
    \label{alg:MIA_defense}
\end{algorithm}


\zhigang{To evaluate the effect of our proposed regularizer on both the underlying model's prediction performance and the resilience to MIA, we setup our experiments as follows.} \shakila{For all the experiments,} \zhigang{the size of training dataset} \shakila{is fixed to 10,000 with <50\% balance in the class. Fairness is measured on 5 features. Number of epochs $k$ is set to $50$ and number of iterations $l$ is set to $5$.} \zhigang{We measure the performance of the selected regularizers in terms of base ML model training and test accuracy, and MIA accuracy.}


\zhigang{Table~\ref{tab:reg_acc} illustrates the results of our evaluations over different choices of the regularizer: no regularizer involved (None), L1 and L2 regularizers (L1 \& L2), the group fairness ($\delta_{g}$), the predictive fairness ($\delta_{p}$), the individual fairness ($\delta_{i}$), and the mutual information ($I_{X}$). In addition, we also record the maximum decrement of our proposed regularizers during the training process (the column Max Diff. in Table~\ref{tab:reg_acc}).}

\zhigang{Particularly,} for all the 
studied \zhigang{regularizers that use fairness values}, the obtained results show success both in terms of improving the model's performance and reducing MIA accuracy compared to the models \zhigang{using either no} 
regularizer or 
L1 and L2 regularizers. \zhigang{However, the mutual information does not provide the promising improvement as we expect. Comparing the total decrement of the regularizers over the training process (Max Diff. in Table~\ref{tab:reg_acc}) with our results in Figure~\ref{fig:p_mi_attack} and Figure~\ref{fig:p_fairness_attack}, the success of our proposed regularizers depends on whether their overall decrement throughout the training process can significantly decrease the MIA accuracy.} 
\gio{In the case of mutual information the total decrement is in the range $[0, 0.01]$, and such a small variation cannot significantly improve MIA resilience as depicted in the results in Figure~\ref{fig:p_mi_attack}. It should also be noted that the effectiveness of regularizers is bound to their initial values, i.e. the ones in the initial steps of the training. In particular, large initial values imply that the model may not gain sufficient improvement in resilience to MIA over the training process, especially when regularizer decrements are small or the training converges very quickly.}


\zhigang{In addition, we also perform the experiments on different regularizer functions}. \shakila{
For example, in case of the Texas dataset, while using $\delta_p$ as a regularizer, modifying the amount of regularizing to $\sum_{j}|(\theta)^2\times r|$ reduces the MIA accuracy to $54\%$ and the train and test accuracy to 83.9\% and 84.8\% respectively.} 


\begin{table}[t]
    \centering
    \captionsetup{justification=centering}
    \subfloat[Purchase dataset.]{
    \centering
    \begin{tabular*}{.47\textwidth}{ccccc}
        Regularizers & Train Acc. &	Test Acc.&	\zhigang{MIA} Acc. & Max Diff.\\\hline
        None & 73.04&	72.80&	73.89& --\\ 
        L1 & 76.78&	76.81&	55.29 & --\\ 
        L2 & 76.69&	76.94&	59.13 &--\\\hline
        $\delta_g$ & 85.00&	84.72&	50.03& 0.66\\
        $\delta_p$ & 84.72& 84.59&48.84& 0.50  \\
        $\delta_i$ & 84.89&	84.93&	51.99&	0.28\\
        $I_X$& 73.14&	73.35&	51.15& 0.02\\ 
        \end{tabular*}
        }
    \hfill
    \subfloat[Texas dataset.]{
    \centering
    \begin{tabular*}{.47\textwidth}{ccccc}
        Regularizers & Train Acc. &	Test Acc. &	\zhigang{MIA} Acc. & Max Diff.\\ \hline
        None & 76.17&	75.87&	74.04& --\\ 
        L1 & 79.93&	80.34&	55.76& -- \\ 
        L2 & 78.63&	78.74&	55.72& --\\
        \hline
        $\delta_g$ & 84.75&84.71&	53.27& 1\\
        $\delta_p$ & 84.73&	84.80&	58.53& 0.35 \\
        $\delta_i$ & 84.71&	84.72&	58.15& 0.33\\
        $I_X$& 70.93&	70.77&	59.40& 0.01\\ 
    \end{tabular*}
    }
    \hfill
    \subfloat[Adult dataset.]{
    \centering
    \begin{tabular*}{.47\textwidth}{ccccc}
        Regularizers & Train Acc. &	Test Acc. &	\zhigang{MIA} Acc. & Max Diff. \\ \hline
        None & 76.54&	76.52&	73.36& --\\ 
        L1 & 76.42&	76.72&	52.41& --\\ 
        L2 & 69.64&	69.83&	53.95& --\\\hline
        $\delta_g$ & 84.49&	84.52&	52.33& 0.8\\
        $\delta_p$ & 84.57&	84.15&	51.84& 0.51 \\
        $\delta_i$ & 84.41&84.60&50.25& 0.48\\
        $I_X$& 70.14&	70.31&	53.34& 0.01\\
    \end{tabular*}
    }
    \caption{Train, test and MIA accuracy (\%) for different regularizers \zhigang{(Max Diff.: the maximum reduction of the regularizer values during the training process. When the training is terminated, those regularizers would end up with their optimal status, which is 0)}.
    }
    \label{tab:reg_acc}
\end{table}

\section{Conclusion \& Future Work} \label{sec:conclusion}


We provided a comprehensive analysis analyzes the effect of several data and model properties \zhigang{on} 
MIA \zhigang{(using the shadow model approach~\cite{10shokri2017membership}) accuracy}.

Our investigation shows that \zhigang{a vulnerable training dataset typically has at least one of the following properties: relatively small (compared to the adversary's shadow dataset), imbalanced classes or features, low entropy.} 
On the other hand, choosing the right model with proper hyperparameter settings \zhigang{(deeper and larger neural networks)} can reduce the vulnerability to MIA. An ML model that learns too much about the training data and produces higher mutual information between the records and the model's parameters is also more likely to exhibit a high MIA accuracy. Additionally, 
we found that fairer ML models are more resilient against MIA.

We further study how our findings on data and model properties versus MIA accuracy can be utilised to strengthen ML privacy. We apply model properties that positively impact the MIA accuracy, such as group, predictive and individual fairness and mutual information between the records and the parameters as regularizers in the ML model. We observed that all the \zhigang{three fairness} 
regularizers reduce the MIA accuracy as well as the model's training loss. 



\textbf{Limitations and Future works.} In this paper, we only considered MIA \zhigang{(in its shadow model approach~\cite{10shokri2017membership})} as a privacy threat to ML models, while other variants of black-box attacks such as \zhigang{the relaxed MIA~\cite{salem2018ml}}, model inversion attacks \cite{veale2018algorithms} and attribute inference attacks \cite{attriguard} are yet to be studied. In addition, it is necessary to study the information leakage risks in other variants of ML algorithms, such as re-enforced ML and agent-based modelling. Besides, as the defenses are model-specific, further research needs to be conducted towards formulating a comprehensive defense mechanism against MIA that is not bounded by the type of the target model. 

While assessing all factors contributing to MIA success certainly remains an open research problems, our results show that many several properties beyond overfitting could be leveraged to further protect ML Models against membership inference attacks without giving away the model's prediction accuracy.   

For future work we plan to :
\begin{itemize}
    \item Further explore other data and model properties and their impact on different black-box and white-box ML adversarial attacks;
    \item Further investigate those properties, such as the number of features, that show weak correlation with MIA or inconsistent results across datasets, and identify other impactful data and model properties;
    \item Develop model-independent defense mechanisms against the attacks, suitable to a wide range of ML algorithms, e.g. applicable to the decision trees or the random forests.
\end{itemize}

\bibliographystyle{plain}
\bibliography{ref.bib}

\begin{thebibliography}{10}

\bibitem{backes2016membership}
Michael Backes, Pascal Berrang, Mathias Humbert, and Praveen Manoharan.
\newblock Membership privacy in microrna-based studies.
\newblock In {\em Proceedings of the 2016 ACM SIGSAC Conference on Computer and
  Communications Security}, pages 319--330. ACM, 2016.

\bibitem{barocas2017fairness}
Solon Barocas, Moritz Hardt, and Arvind Narayanan.
\newblock Fairness in machine learning.
\newblock {\em NIPS Tutorial}, 2017.

\bibitem{bassily2014private}
Raef Bassily, Adam Smith, and Abhradeep Thakurta.
\newblock Private empirical risk minimization: Efficient algorithms and tight
  error bounds.
\newblock In {\em 2014 IEEE 55th Annual Symposium on Foundations of Computer
  Science}, pages 464--473. IEEE, 2014.

\bibitem{binns2017fairness}
Reuben Binns.
\newblock Fairness in machine learning: Lessons from political philosophy.
\newblock {\em arXiv preprint arXiv:1712.03586}, 2017.

\bibitem{chen2019gan}
Dingfan Chen, Ning Yu, Yang Zhang, and Mario Fritz.
\newblock Gan-leaks: A taxonomy of membership inference attacks against gans.
\newblock {\em arXiv preprint arXiv:1909.03935}, 2019.

\bibitem{lasagne_2015}
Lasagne contributors.
\newblock Welcome to lasagne - lasagne 0.2.dev1 documentation.
\newblock \url{https://lasagne.readthedocs.io/en/latest/}, 2015.
\newblock Accessed: 2020-05-01.

\bibitem{texas}
Dshs.texas.gov.
\newblock Hospital discharge data public use data file.
\newblock \url{https://www.dshs.texas.gov/THCIC/Hospitals/Download.shtm}, 2019.
\newblock {Accessed: 2019-08-30}.

\bibitem{dwork2012fairness}
Cynthia Dwork, Moritz Hardt, Toniann Pitassi, Omer Reingold, and Richard Zemel.
\newblock Fairness through awareness.
\newblock In {\em Proceedings of the 3rd innovations in theoretical computer
  science conference}, pages 214--226. ACM, 2012.

\bibitem{entropy_2019}
Wikimedia Foundation.
\newblock Entropy (information theory).
\newblock \url{https://en.wikipedia.org/wiki/Entropy_(information_theory)}, Nov
  2019.
\newblock {Accessed: 2019-11-27}.

\bibitem{gajane2017formalizing}
Pratik Gajane and Mykola Pechenizkiy.
\newblock On formalizing fairness in prediction with machine learning.
\newblock {\em arXiv preprint arXiv:1710.03184}, 2017.

\bibitem{17hayes2019logan}
Jamie Hayes, Luca Melis, George Danezis, and Emiliano De~Cristofaro.
\newblock Logan: Membership inference attacks against generative models.
\newblock {\em Proceedings on Privacy Enhancing Technologies},
  2019(1):133--152, 2019.

\bibitem{hilprecht2019monte}
Benjamin Hilprecht, Martin H{\"a}rterich, and Daniel Bernau.
\newblock Monte carlo and reconstruction membership inference attacks against
  generative models.
\newblock {\em Proceedings on Privacy Enhancing Technologies},
  2019(4):232--249, 2019.

\bibitem{hisamoto2019membership}
Sorami Hisamoto, Matt Post, and Kevin Duh.
\newblock Membership inference attacks on sequence-to-sequence models.
\newblock {\em arXiv preprint arXiv:1904.05506}, 2019.

\bibitem{irolla2019demystifying}
Paul Irolla and Gr{\'e}gory Ch{\^a}tel.
\newblock Demystifying the membership inference attack.
\newblock In {\em 2019 12th CMI Conference on Cybersecurity and Privacy (CMI)},
  pages 1--7. IEEE, 2019.

\bibitem{attriguard}
Jinyuan Jia and Neil~Zhenqiang Gong.
\newblock Attriguard: {A} practical defense against attribute inference attacks
  via adversarial machine learning.
\newblock {\em CoRR}, abs/1805.04810, 2018.

\bibitem{kaggle}
Kaggle.
\newblock Acquire valued shoppers challenge.
\newblock
  \url{https://www.kaggle.com/c/acquire-valued-shoppers-challenge/data}, 2014.
\newblock {Accessed: 2019-08-30}.

\bibitem{cifar_2009}
Alex Krizhevsky.
\newblock Cifar-10 and cifar-100 datasets.
\newblock \url{https://www.cs.toronto.edu/~kriz/cifar.html}, 2009.
\newblock Accessed: 2020-05-1.

\bibitem{theano_2008}
LISA Lab.
\newblock Welcome - theano 1.0.0 documentation.
\newblock \url{http://deeplearning.net/software/theano/}, 2008.
\newblock Accessed: 2020-05-01.

\bibitem{liu2019socinf}
Gaoyang Liu, Chen Wang, Kai Peng, Haojun Huang, Yutong Li, and Wenqing Cheng.
\newblock Socinf: Membership inference attacks on social media health data with
  machine learning.
\newblock {\em IEEE Transactions on Computational Social Systems}, 2019.

\bibitem{long2018understanding}
Yunhui Long, Vincent Bindschaedler, Lei Wang, Diyue Bu, Xiaofeng Wang, Haixu
  Tang, Carl~A Gunter, and Kai Chen.
\newblock Understanding membership inferences on well-generalized learning
  models.
\newblock {\em arXiv preprint arXiv:1802.04889}, 2018.

\bibitem{21mcsherry2009differentially}
Frank McSherry and Ilya Mironov.
\newblock Differentially private recommender systems: Building privacy into the
  netflix prize contenders.
\newblock In {\em Proceedings of the 15th ACM SIGKDD international conference
  on Knowledge discovery and data mining}, pages 627--636. ACM, 2009.

\bibitem{nasr2018machine}
Milad Nasr, Reza Shokri, and Amir Houmansadr.
\newblock Machine learning with membership privacy using adversarial
  regularization.
\newblock In {\em Proceedings of the 2018 ACM SIGSAC Conference on Computer and
  Communications Security}, pages 634--646. ACM, 2018.

\bibitem{pyrgelis2017knock}
Apostolos Pyrgelis, Carmela Troncoso, and Emiliano De~Cristofaro.
\newblock Knock knock, who's there? membership inference on aggregate location
  data.
\newblock {\em arXiv preprint arXiv:1708.06145}, 2017.

\bibitem{12rahman2018membership}
Md~Atiqur Rahman, Tanzila Rahman, Robert Laganiere, Noman Mohammed, and Yang
  Wang.
\newblock Membership inference attack against differentially private deep
  learning model.
\newblock {\em Transactions on Data Privacy}, 11(1):61--79, 2018.

\bibitem{uci_adult}
UCI Machine~Learning Repository.
\newblock Adult data set.
\newblock \url{https://archive.ics.uci.edu/ml/datasets/Adult}, 1996.
\newblock Accessed: 2019-08-30.

\bibitem{reynolds2009gaussian}
Douglas~A Reynolds.
\newblock Gaussian mixture models.
\newblock {\em Encyclopedia of biometrics}, 741, 2009.

\bibitem{salem2018ml}
Ahmed Salem, Yang Zhang, Mathias Humbert, Mario Fritz, and Michael Backes.
\newblock Ml-leaks: Model and data independent membership inference attacks and
  defenses on machine learning models.
\newblock {\em arXiv preprint arXiv:1806.01246}, 2018.

\bibitem{schonherr2018adversarial}
Lea Schonherr, Katharina Kohls, Steffen Zeiler, Thorsten Holz, and Dorothea
  Kolossa.
\newblock Adversarial attacks against automatic speech recognition systems via
  psychoacoustic hiding.
\newblock {\em arXiv preprint arXiv:1808.05665}, 2018.

\bibitem{gauss_2007}
scikit-learn developers.
\newblock 2.1. gaussian mixture models.
\newblock \url{https://scikit-learn.org/stable/modules/mixture.html}, 2007.
\newblock Accessed: 2020-05-1.

\bibitem{10shokri2017membership}
Reza Shokri, Marco Stronati, Congzheng Song, and Vitaly Shmatikov.
\newblock Membership inference attacks against machine learning models.
\newblock In {\em Security and Privacy (SP), 2017 IEEE Symposium on}, pages
  3--18. IEEE, 2017.

\bibitem{19srivastava2014dropout}
Nitish Srivastava, Geoffrey Hinton, Alex Krizhevsky, Ilya Sutskever, and Ruslan
  Salakhutdinov.
\newblock Dropout: a simple way to prevent neural networks from overfitting.
\newblock {\em The Journal of Machine Learning Research}, 15(1):1929--1958,
  2014.

\bibitem{truex2019effects}
Stacey Truex, Ling Liu, Mehmet~Emre Gursoy, Wenqi Wei, and Lei Yu.
\newblock Effects of differential privacy and data skewness on membership
  inference vulnerability.
\newblock {\em arXiv preprint arXiv:1911.09777}, 2019.

\bibitem{18truex2018towards}
Stacey Truex, Ling Liu, Mehmet~Emre Gursoy, Lei Yu, and Wenqi Wei.
\newblock Towards demystifying membership inference attacks.
\newblock {\em arXiv preprint arXiv:1807.09173}, 2018.

\bibitem{veale2018algorithms}
Michael Veale, Reuben Binns, and Lilian Edwards.
\newblock Algorithms that remember: model inversion attacks and data protection
  law.
\newblock {\em Philosophical Transactions of the Royal Society A: Mathematical,
  Physical and Engineering Sciences}, 376(2133):20180083, 2018.

\bibitem{verma2018fairness}
Sahil Verma and Julia Rubin.
\newblock Fairness definitions explained.
\newblock In {\em 2018 IEEE/ACM International Workshop on Software Fairness
  (FairWare)}, pages 1--7. IEEE, 2018.

\bibitem{yaghini2019disparate}
Mohammad Yaghini, Bogdan Kulynych, and Carmela Troncoso.
\newblock Disparate vulnerability: on the unfairness of privacy attacks against
  machine learning.
\newblock {\em arXiv preprint arXiv:1906.00389}, 2019.

\bibitem{yeom2018privacy}
Samuel Yeom, Irene Giacomelli, Matt Fredrikson, and Somesh Jha.
\newblock Privacy risk in machine learning: Analyzing the connection to
  overfitting.
\newblock In {\em 2018 IEEE 31st Computer Security Foundations Symposium
  (CSF)}, pages 268--282. IEEE, 2018.

\bibitem{zhao2019inferring}
Benjamin Zi~Hao Zhao, Hassan~Jameel Asghar, Raghav Bhaskar, and Mohamed~Ali
  Kaafar.
\newblock On inferring training data attributes in machine learning models.
\newblock {\em arXiv preprint arXiv:1908.10558}, 2019.

\end{thebibliography}

\end{document}